\pdfoutput=1

\documentclass[11pt]{article}

\usepackage{acl}

\usepackage{times}
\usepackage{latexsym}
\usepackage{amssymb}
\usepackage{multirow}
\usepackage{booktabs}
\usepackage{hyperref}
\usepackage{dsfont}
\usepackage{amsmath}
\usepackage{caption}
\usepackage{subcaption}
\usepackage{todonotes}
\usepackage{siunitx}
\usepackage[T1]{fontenc}
\usepackage[utf8]{inputenc}
\usepackage{CJKutf8}
\usepackage{graphicx}
\graphicspath{ {./figures/} }
\usepackage{microtype}
\usepackage{xcolor}
\usepackage{arydshln}
\usepackage{amsfonts,dcolumn}

\usepackage{bm}

\def\eqref#1{equation~\ref{#1}}

\def\1{\bm{1}}

\def\vz{{\bm{z}}}

\DeclareMathAlphabet{\mathsfit}{\encodingdefault}{\sfdefault}{m}{sl}
\SetMathAlphabet{\mathsfit}{bold}{\encodingdefault}{\sfdefault}{bx}{n}

\usepackage{array}
\usepackage{xspace}

\newcolumntype{L}[1]{>{\raggedright\let\newline\\\arraybackslash\hspace{0pt}}m{#1}}
\newcolumntype{C}[1]{>{\centering\let\newline\\\arraybackslash\hspace{0pt}}m{#1}}
\newcolumntype{R}[1]{>{\raggedleft\let\newline\\\arraybackslash\hspace{0pt}}m{#1}}

\makeatletter
\def\adl@drawiv#1#2#3{%
        \hskip.5\tabcolsep
        \xleaders#3{#2.5\@tempdimb #1{1}#2.5\@tempdimb}%
                #2\z@ plus1fil minus1fil\relax
        \hskip.5\tabcolsep}
\newcommand{\cdashlinelr}[1]{%
  \noalign{\vskip\aboverulesep
           \global\let\@dashdrawstore\adl@draw
           \global\let\adl@draw\adl@drawiv}
  \cdashline{#1}
  \noalign{\global\let\adl@draw\@dashdrawstore
           \vskip\belowrulesep}}
\makeatother

\makeatletter
\DeclareRobustCommand\onedot{\futurelet\@let@token\@onedot}
\def\@onedot{\ifx\@let@token.\else.\null\fi\xspace}

\def\eg{e.g\onedot,\xspace} 
\def\ie{i.e\onedot,\xspace}

\makeatother

\newcommand{\modelfine}{\textsc{Finetune}\xspace}
\newcommand{\modelfinedeb}{\textsc{Finetune}\textsc{-debias}\xspace}
\newcommand{\modeladapter}{\textsc{Adapter}\xspace}
\newcommand{\modeladapterdeb}{\textsc{Adapter}\textsc{-debias}\xspace}
\newcommand{\modeldiff}{\textsc{DiffPrun}\xspace}
\newcommand{\modeldiffdeb}{\textsc{DiffPrun}\textsc{-debias}\xspace}
\newcommand{\modelmodular}{\textsc{MoDDiffy}\xspace}
\newcommand{\modelmodularstar}{\textsc{MoDDiffy}\textsc{-*}\xspace}
\newcommand{\modelmodularpar}{\textsc{MoDDiffy}\textsc{-par}\xspace}
\newcommand{\modelmodularseq}{\textsc{MoDDiffy}\textsc{-post}\xspace}
\newcommand{\modelinlp}{\textsc{INLP}\xspace}

\title{Modular and On-demand Bias Mitigation with Attribute-Removal Subnetworks}

\author{Lukas Hauzenberger,~~Shahed Masoudian,~~Deepak Kumar,\\{\bf Markus Schedl,}~~{\bf Navid Rekabsaz} \\
  Johannes Kepler University Linz, Austria\\
  Linz Institute of Technology, AI Lab\\
  \texttt{\{first\_name.family\_name\}@jku.at}  
}

\begin{document}
\maketitle
\begin{abstract}
Societal biases are reflected in large pre-trained language models and their fine-tuned versions on downstream tasks. Common in-processing bias mitigation approaches, such as adversarial training and mutual information removal, introduce additional optimization criteria, and update the model to reach a new debiased state. However, in practice, end-users and practitioners might prefer to switch back to the original model, or apply debiasing only on a specific subset of protected attributes. To enable this, we propose a novel modular bias mitigation approach, consisting of stand-alone highly sparse debiasing subnetworks, where each debiasing module can be integrated into the core model on-demand at inference time. Our approach draws from the concept of \emph{diff} pruning, and proposes a novel training regime adaptable to various representation disentanglement optimizations. We conduct experiments on three classification tasks with gender, race, and age as protected attributes. The results show that our modular approach, while maintaining task performance, improves (or at least remains on-par with) the effectiveness of bias mitigation in comparison with baseline finetuning. Particularly on a two-attribute dataset, our approach with separately learned debiasing subnetworks shows effective utilization of either or both the subnetworks for selective bias mitigation. 

\end{abstract}

\section{Introduction}
\label{sec:introduction}


A large body of research evidences the existence of societal biases and stereotypes in pre-trained language models (PLMs)~\cite{zhao2019gender,sheng2019woman,rekabsaz_2021_societal}, and their potential harms when used in down-stream tasks~\cite{blodgett2020language,dearteaga_2019_bios,rekabsaz_2020_neural,stanovsky2019evaluating}. Common in-processing approaches to bias mitigation update a model's (typically all) parameters to satisfy specific attribute erasure criteria through optimization methods such as adversarial training~\cite{elazar_2018_adv,rekabsaz_2021_societal}, and mutual information reduction~\cite{colombo_2021_novel}. These methods are shown to be effective in reducing the footprint of protected attributes (\eg gender, race, etc.) in the resulting model. 

However when using such debiasing models in practice and in specific use-cases, system designers or end-users might still prefer to instead use the original model, or a debiased variation in respect to a particular subset of protected attributes. This can be due to various reasons such as the nature of a given input, preference of an individual end-user, or fairness-utility trade-off considerations. For instance, while a bias-aware model should indeed be agnostic to genders when the input is about gender-neutral occupations (such as \emph{nurse} or \emph{CEO}), certain topics like \emph{pregnancy} may specifically require gender information for a correct model decision.\footnote{See also the discussion in \citet{krieg2022grep} about the need to separate bias-sensitive queries from ``normal'' ones.} Also as shown in previous studies~\cite{zerveas_2022_mitigating,biega2018equity,rekabsaz_2021_societal}, since improving fairness on specific tasks may come with the cost of performance degradation, it is necessary to provide on-demand control over whether to impose fairness/debiasing criteria. Using existing approaches, this would require maintaining and deploying multiple large parallel models for every protected attribute, resulting in overly complex and resource-heavy pipelines and increased latency.

To address this, we introduce a novel modular bias mitigation approach using sparse weight-difference networks. In our approach, the required changes in a model's parameters for erasing a bias attribute are stored in a decoupled subnetwork, trained simultaneously by a debiasing and a sparsification objective. At inference time, adding each debiasing module to the core model results in delivering debiasing qualities to a model's prediction in respect to the corresponding protected attribute. Our approach extends the principle idea of \emph{diff pruning}~\cite{guo_2021_diffpruning} introduced for parameter-efficient task training to bias mitigation, by viewing the objective of erasing a protected attribute as a stand-alone \emph{diff} module. This module replaces fine-tuning by training only a small set of parameters added to the corresponding PLM's parameters, to deliver bias mitigation of a specific protected attribute. We further propose a novel procedure to train such debiasing subnetworks separately, and to selectively add an arbitrary set of them to the core model at inference time (\S\ref{sec:method}). 

Our approach can be applied to any debiasing and representation disentanglement method, provided that its objective has separate learning signals for the task and each protected attribute. In comparison with adapter networks~\cite{rebuffi2017learning,houlsby_2019_param_efficient}, as shown by \citet{guo_2021_diffpruning} and also evidenced in our experiments, even more parameter-efficiency can be provided. Additionally, since our approach extends the base model with \emph{diff} subnetworks, the resulting model is expected to perform (at least) as good as the fine-tuning variation, avoiding possible performance degradations. The modularity of our approach supports separating the process of developing debiasing solutions for a task from using them, such that stand-alone debiasing modules can be created and shared, and later be utilized in a final system on-demand. 

\begin{figure*}[t]
  \centering
  \includegraphics[width=\textwidth]{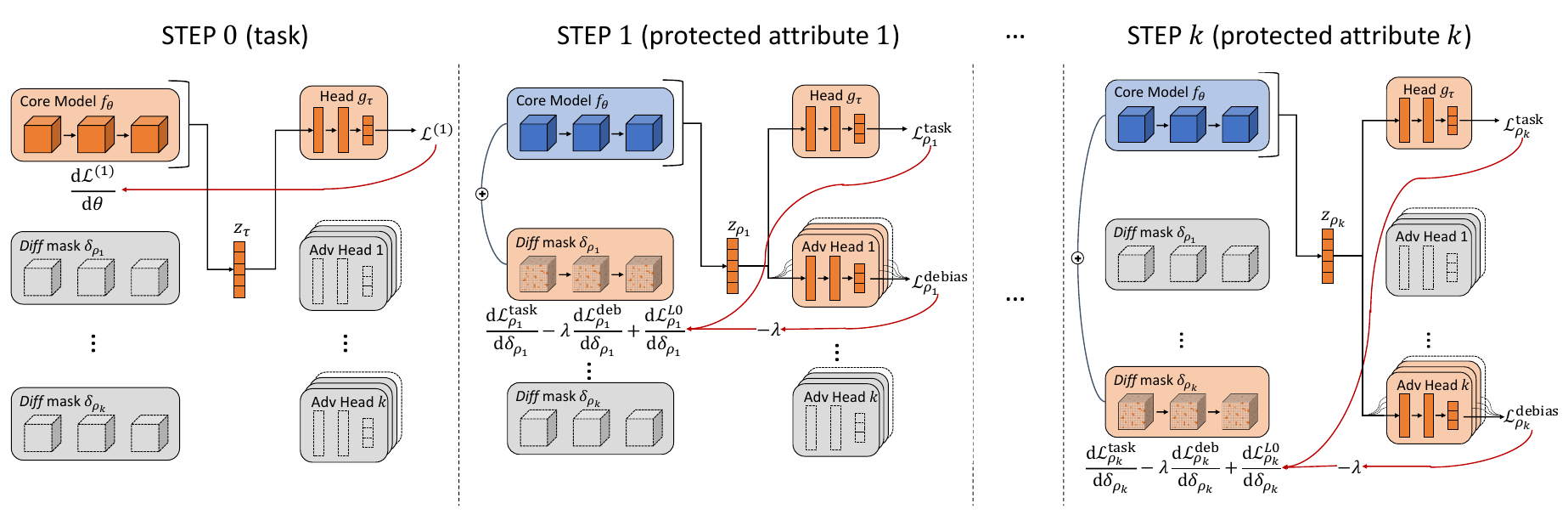}
  \caption{Training procedure of \modelmodularpar in one batch with adversarial bias removal for $k$ protected attributes. The color orange indicates the trainable parameters during each step, while blue shows the frozen ones. The core model learns the task in Step 0, and the subnetworks --- each responsible for debiasing a protected attribute --- are trained in the next steps. In \modelmodularseq training regime, after finetuning the core model in Step 0, the task head parameters (together with the ones of the core model) remain frozen in the next steps.}
  \label{fig:modular_multi1}
\end{figure*}

We evaluate our approach on three bias mitigation tasks: occupation prediction from biographies involving gender~\cite{dearteaga_2019_bios}, hate speech detection with dialect-based race as sensitive attribute~\cite{founta_2018_twitter}; and mention prediction in tweets with two attributes of gender and age of the authors~\cite{Pardo2016OverviewOT}. The last dataset particularly enables the study of combining independently trained debiasing modules (details in \S\ref{sec:setup}). The evaluation results show that our approach, due to learning the subnetworks specialized on the narrow functionality of debiasing an attribute, provides on par or better debiasing performance in comparison with strong baselines. Additionally, we observe that on the mention detection task, learning the debiasing subnetworks post-hoc to model training provides effective results when combining the two (independently trained) subnetworks at inference time. Remarkably, these results are achieved with debiasing subnetworks of maximum 1\% size of the core model (BERT-Base), in some cases (e.g., for gender attribute) even only 0.01\% (details in \S\ref{sec:results}). 


\section{Related Work}
\label{sec:related}

\subsection{Parameter-efficient and Modular Training}

The concept of subnetworks is grounded in the lottery ticket hypothesis~\cite{lottery_ticket_original_Frankle2019}, stating that in deep neural networks, one can find many sparse subnetworks with a capacity comparable to that of the base network. \citet{zhou2019deconstructing} show that spotting such subnetworks through binary masks can in fact be seen as a form of model training. \citet{zhao_2020_masking} further consider the magnitude of the parameters for pruning and filter out the ones lower than a specific threshold. \citet{guo_2021_diffpruning} use $L_{0}$-regularization~\cite{louizos_2018_l0} to reduce the number of active neurons. \citet{hu_2021_lora} propose low-rank adaptation via rank decomposition matrices. These methods are extended by structural pruning approaches with additional architectural constraints, commonly led to a higher parameter-efficiency~\cite{lagunas2021block,jaszczur2021sparse}.

Recent studies exploit subnetworks in the context of multi-task learning. \citet{wortsman2020supermasks} isolate the learning signal of each task in a separate masking subnetwork, while \citet{zaken_2021_bitfit} only finetune the bias weights. \citet{xu_2021_child} learn a subset of parameters via masking out the gradients of other parameters during backward-pass. \citet{guo_2021_diffpruning} suggest learning a sparse \emph{diff} subnetwork for each task, whose parameter values are added to the corresponding parameters of the base network. An alternative architecture are adapter networks, first introduced in the context of multi-task learning~\cite{rebuffi2017learning,houlsby_2019_param_efficient,pmlr-v97-stickland19a}, and are then extended in respect to their parameter efficiency~\cite{ruckle2021adapterdrop,han2021robust}, architectural variations~\cite{DBLP:conf/nips/MahabadiHR21}, and transfer learning capacity~\cite{pfeiffer2021adapterfusion}. As stated by \cite{sung_2021_fisher}, adapters (in their original form) slightly increase the inference cost of a model in comparison with the original form or pruning-based variations. Our proposed approach contributes to this line of research by extending this concept to on-demand bias mitigation.



\subsection{Fairness \& Bias Mitigation in NLP}

Several studies explore methods for debiasing PLMs, such as linearly projecting embeddings into the space with minimum correlations to protected features~\cite{ravfogel_2020_null,kaneko_2021_debiasing,bolukbasi_2016_bias_embedding}, utilizing a distribution alignment loss~\cite{guo_2022_auto}, or penalizing bias by utilizing the encoded information in models~\cite{schick_2021_self}. Adversarial training, originally introduced in domain adaptation \cite{ganin_2016_domain_adv, ganin_2015_domainadaptation} is utilized in the context of fair representation learning~\cite{xie_2017_controllable,madras_2018_learning}, and later to erase demographic data from text classifiers~\cite{elazar_2018_adv, barrett_2019_adversarial,han_2021_diverse,wang_2021_dynamically}, information retrieval models~\cite{rekabsaz_2021_societal}, and recommendation systems~\cite{ganhoer2022mitigating}. Mutual information removal is an alternative approach, which minimizes the approximate upper bound of the common information between the task and protected attributes~\cite{cheng_2020_improving,colombo_2021_novel}. Our work utilizes these optimizations to learn a novel modular on-demand bias mitigation approach.



Few recent studies explore parameter-efficient training for bias mitigation. \citet{lauscher_2021_sustainable} approach debiasing PLMs using a stack of adapters. While shown effective in practice, the adapters in the higher levels inherently depend on the ones in the lower levels and cannot be learned nor utilized stand-alone. \citet{zhang_2021_disentangling} approach bias mitigation with binary masks applied to the base network. More recently and in the context of removing spurious shortcuts in natural language understanding datasets, \citet{meissner_2022_debias} train sparse binary masks on a finetuned model. Our work extends this line of research by encapsulating concept erasure modules in separate \emph{diff} subnetworks for each protected attribute, and selectively applying them to the base model at inference time.



\section{Modular Debiasing with \emph{Diff} Subnets}
\label{sec:method}
We start with defining the general approach to model bias mitigation. We consider an arbitrary PLM denoted by $f_{\boldsymbol{\theta}}$ with the set of parameters $\boldsymbol{\theta}$. The model learns the task $\tau$ using the loss function $\mathcal{L}_{\tau}$. The predictions of $f_{\boldsymbol{\theta}}$ might be sensitive to the variations in any of the $k$ protected attributes of the set $\mathrm{P}=\{\rho_1,...,\rho_k\}$. The bias mitigation objective is to make the model invariant to these variations while maintaining the effectiveness on the task, approached by defining the debiasing loss $\mathcal{L}_{\rho_{i}}$ for the protected attribute $\rho_{i}$. We discuss two realizations of this loss function in Section~\ref{sec:method:objectives}. A debiased model in respect to attributes $\mathrm{P}$ is achieved by training on the following loss function:
\begin{equation} 
\mathcal{L}_{total} = \mathcal{L}_{\tau} + \mathcal{L}_{\rho_{1}} + ... + \mathcal{L}_{\rho_{k}}
\label{eq:method:generalloss}
\end{equation}
Using $\mathcal{L}_{total}$, one can finetune all parameters of the model~\cite{elazar_2018_adv,rekabsaz_2021_societal}, or utilize any parameter-efficient training such as adapters~\cite{lauscher_2021_sustainable}, \emph{diff} pruning~\cite{guo_2021_diffpruning}, or binary masks~\cite{zhang_2021_disentangling}. We use some of these methods as baselines, explained in the following sections. In the remainder of this section, we introduce our \emph{Modular Debiasing with Diff Subnetworks} (\modelmodular), explain its training and inference procedure, and describe two debiasing optimization methods. 

\subsection{\modelmodular}
We aim to encapsulate the debiasing functionality of the protected attribute $\rho_{i}$, provided by the signal of the corresponding loss $\mathcal{L}_{\rho_{i}}$, into the sparse \emph{diff} subnetwork characterized by the set of parameters $\boldsymbol{\delta}_{\rho_{i}}$. Each parameter in $\boldsymbol{\delta}_{\rho_{i}}$ corresponds to a parameter in $\boldsymbol{\theta}$, such that adding $\boldsymbol{\delta}_{\rho_{i}}$ to the corresponding parameters in $\boldsymbol{\theta}$ results in debiasing the model $f$. 

To learn the model, let us consider a training data item in form of $\langle x,y_{\tau},y_{\rho_{1}},...,,y_{\rho_{k}}\rangle$, where $x$ is the input, $y_{\tau}$ the task label, and $y_{\rho_{i}}$ denotes the label of the corresponding protected attribute $\rho_{i}$. Figure~\ref{fig:modular_multi1} depicts the \modelmodular approach with adversarial bias mitigation optimization. We first optimize for the task by encoding $x$ into the vector $\vz_{\tau}$ using $f(.;{\boldsymbol{\theta}})$. An arbitrary decoder network denoted as $g_{\tau}$ uses $\vz_{\tau}$ to predict the task output, and task loss is calculated using cross entropy (CE), as formulated below:
\begin{equation}
\vz_{\tau}=f(x;\boldsymbol{\theta}),\quad \mathcal{L}^{(0)}_{total} = \text{CE}(g_{\tau}(\vz_{\tau}), y_{\tau})
\end{equation}

This loss updates $\boldsymbol{\theta}$ as well as the  parameters of $g_{\tau}$. While in our experiments we opt for fully finetuning $f_{\theta}$, in practice the model can be trained using any parameter-efficient method.


Next, we iterate over the protected attributes, and in each step learn the corresponding debiasing \emph{diff} subnetwork. Specifically in step $i$ dedicated to the protected attribute $\rho_i$, we learn the set of sparse additional parameters $\boldsymbol{\delta}_{\rho_{i}}$ such that by being added to $\boldsymbol{\theta}$, the resulting $f(.;\boldsymbol{\theta}+\boldsymbol{\delta}_{\rho_{i}})$ model is debiased. Learning $\boldsymbol{\delta}_{\rho_{i}}$ is characterized by three loss functions. The first loss, $\mathcal{L}^{\text{task}}_{\rho_i}$ maintains the task performance when the task output is predicted from the altered encoder formulated below:
\begin{equation}
\vz_{\rho_{i}}=f(x;\boldsymbol{\theta}+\boldsymbol{\delta}_{\rho_{i}}),\quad \mathcal{L}^{\text{task}}_{\rho_i} = \text{CE}(g_{\tau}(\vz_{\rho_{i}}), y_{\tau})
\end{equation}

The second is the debiasing loss $\mathcal{L}_{\rho_i}^{\text{debias}}$, defined based on $\vz_{\rho_{i}}$ and the label of the corresponding protected attribute $y_{\rho_{i}}$. We discuss adversarial bias removal and mutual information reduction as two possible realization of this representation disentanglement loss in Section~\ref{sec:method:objectives}. The third loss imposes the sparsity constraint, defined as the $L_0$ regularization of $\boldsymbol{\delta}_{\rho_i}$. The $L_0$ loss aims to reduce the number of non-zero parameters, namely the term $\sum_{j=1}^{\left|\boldsymbol{\delta}_{\rho_i}\right|} \mathds{1}\{\delta_{\rho_i,j} \neq 0\}$, and is realized with the differentiable approximation proposed by~\citet{louizos_2018_l0}. Following \citet{guo_2021_diffpruning}, $\boldsymbol{\delta}_{\rho_i}$ is decomposed into the element-wise multiplication of two sets of parameters: $\boldsymbol{\delta}_{\rho_i} = \boldsymbol{m}_{\rho_i} \odot \boldsymbol{w}_{\rho_i}$. The parameter set $\boldsymbol{w}_{\rho_i}$ stores the magnitude changes (\emph{diff} values), and $\boldsymbol{m}_{\rho_i}$ learns to mask out the parameters. $\boldsymbol{m}_{\rho_i}$ is characterized by the hard concrete distribution~\cite{guo_2021_diffpruning} with $(\log{\boldsymbol{\alpha}_{\rho_i}}, 1)$ parameters, and $\gamma<0$ and $\zeta>1$ hyperparameters. This results in the following formulation:
\begin{equation}
  \mathcal{L}_{\rho_{i}}^{L_{0}} = \sum_{j=1}^{\left|\boldsymbol{\delta}_{\rho_i}\right|}{\sigma \left( \log{\alpha_{\rho_i,j}} - \log{(-\frac{\gamma}{\zeta}}) \right)}   
\end{equation}
where $\sigma$ denotes the sigmoid function, and as in \citet{guo_2021_diffpruning} $\beta$ is set to $1$. The masking network can be simply reduced by assigning a mask to a group of parameters (such as a weight matrix, or a layer) instead of each individual one. 



Putting all together, the objective of \modelmodular at step $i>0$ is defined as: 
\begin{equation} 
\mathcal{L}^{(i)}_{total} = \mathcal{L}^{\text{task}}_{\rho_i} + \mathcal{L}_{\rho_i}^{\text{debias}} + \mathcal{L}_{\rho_{i}}^{L_{0}}
\label{eq:method:loss}
\end{equation}

We should note that while each subnetwork is trained independently from the others, the output embeddings $\vz_{\rho_i}$ are passed to the same decoder network $g_{\tau}$. This design choice forces the learned embeddings to remain in the same embedding space, making it possible to add multiple (independently trained) subnetworks together to the core model. Another aspect is that the sparsity rate of each resulting subnetwork is not fixed and may vary due to the factors such as the hyperparameter setting, and the extent of encoded information content. 
To achieve a fixed sparsity rate, we further apply magnitude pruning by only keeping a fixed portion of the parameters with the largest absolute values, and finetuning the resulting parameters with the task and debiasing loss terms. 

We conduct optimization with $\mathcal{L}_{total}$ under two training regimes. In the first one referred to as \modelmodularpar, we repeat the mentioned training steps for each training batch. The second approach referred to as \modelmodularseq trains debiasing subnetworks post-hoc to training the core model. While \modelmodularpar accommodates for a higher flexibility in optimization by training the networks in parallel, \modelmodularseq provides the practical benefits of learning various debiasing solutions for an already trained core model.




Finally at \emph{inference time}, one can use the core model in its original form $f(x;\boldsymbol{\theta})$, or in combination with any of the debiasing subnetworks in the form of $f(x;\boldsymbol{\theta}+\boldsymbol{\delta}_{\rho_{i}})$. Our approach also enables simultaneously debiasing all or any subset of the protected attributes by adding their corresponding subnetworks to the core network, for instance as in 
$f(x;\boldsymbol{\theta}+\boldsymbol{\delta}_{\rho_{1}}+...+\boldsymbol{\delta}_{\rho_{k}})$. We should note that, although these subnetworks reside in the same distributional space, they might affect each others functionality, depending on the inherent nature of the biases as well as the possible correlations between the protected attributes. We will examine this case in the next sections on a dataset with two protected attributes of gender and age.

\subsection{Bias Mitigation Objectives}
\label{sec:method:objectives}
We explain two bias mitigation optimization methods used in \modelmodular in the following. 

\paragraph{Adversarial Bias Removal} This method first defines a new classification head $h_{\rho_{i}}$ for each protected attribute $\rho_i$. This head receives $\vz_{\rho_{i}}$ as input, predicts the corresponding protected attribute, and calculates the cross entropy loss function $\mathcal{L}_{\rho_i}^{\text{debias}}$. This loss needs to remove the information of $\rho_{i}$ from $f$ but train $h_{\rho_{i}}$ such that it can effectively predict the protected attribute. This optimization forms the min-max game: $\mathcal{L}_{\rho_i}^{\text{debias}}=\min_{f}\max_{h_{\rho_i}} \text{CE}(h_{\rho_{i}}(\vz_{\rho_{i}}), y_{\rho_{i}})$. A common approach to turn this loss into a minimization problem is by using gradient resversal layer (GRL)~\cite{ganin_2015_domainadaptation} added before the debiasing heads. GRL multiplies the gradient of $\mathcal{L}_{\rho_i}^{\text{debias}}$ with a factor of $-\lambda_{i}$, and thereby simplifies the learning process to a standard gradient-based optimization, formulated below:
\begin{equation*}
\mathcal{L}_{\rho_i}^{\text{debias}}=\min_{f,h_{\rho_i}} \text{CE}(h_{\rho_{i}}(\vz_{\rho_{i}}), y_{\rho_{i}})
\end{equation*}

\paragraph{Mutual Information (MI) Reduction}
This approach represents a family of algorithms that aims to remove the mutual information of the encoded embeddings of the task and protected attributes. Maximum Mean Discrepancy (MMD), first introduced in the context of domain adaptation~\cite{gretton:mmd,tzeng2014deep}, offers a realization of MI reduction by minimizing the ability to separate the subsets belonging to two protected attributes. In particular, given a set of data points $X$ split into two subsets $X_{\rho_i}^{A}$ and $X_{\rho_i}^{B}$ according to the values of the (binary) protected attribute $\rho_i$, MMD minimizes the distance between the encoded embeddings of the subgroups with the following loss formulation:  
\begin{equation*}
\begin{split}
&\mathcal{L}_{\rho_i}^{\text{debias}}=\\& 
\left(\frac{\sum_{x^A \in X_{\rho_i}^{A}} \phi(f(x^A))}{|X_{\rho_i}^{A}|} - \frac{\sum_{x^B \in X_{\rho_i}^{B}}\phi(f(x^B))}{|X_{\rho_i}^{B}|} \right)^2
\end{split}
\end{equation*}
where $\phi$ is the feature map kernel defined as a linear combination of multiple Gaussian kernels.



\section{Experiment Design}
\label{sec:setup}

\begin{table*}[t]
\centering
    \begin{tabular}{l cccc | cccc}
        \toprule
        \multirow{3}{*}{Model} & \multicolumn{4}{c|}{\textbf{BIOS (gender)}} & \multicolumn{4}{c}{\textbf{FCDL18 (race-dialect)}} \\
         & \multicolumn{2}{c}{Adversarial} & \multicolumn{2}{c|}{MI Reduction} & \multicolumn{2}{c}{Adversarial} & \multicolumn{2}{c}{MI Reduction} \\
        & Task$\uparrow$ & Probe$\downarrow$ & Task$\uparrow$ & Probe$\downarrow$ & Task$\uparrow$ & Probe$\downarrow$ & Task$\uparrow$ & Probe$\downarrow$ \\
        \midrule
        \modelfine & {$84.1_{0.3}$} & {$67.2_{0.7}$} & {$84.1_{0.3}$} & {$67.2_{0.7}$}  & {$82.0_{0.4}$} & {$93.0_{0.4}$} & {$82.0_{0.4}$} & {$93.0_{0.4}$} \\
        \modeladapter& {$84.4_{0.1}$} & {$65.9_{0.1}$} & {$84.4_{0.1}$} & {$65.9_{0.1}$}  & {$80.9_{0.1}$} & {$82.1_{4.1}$} & {$80.9_{0.1}$} & {$82.1_{4.1}$} \\
        \modelfinedeb & {$84.2_{0.1}$} & {$56.9_{0.9}$} & {$84.1_{0.7}$} & {$61.7_{0.8}$} & {$81.9_{0.7}$} & {$84.5_{3.9}$} & {$81.6_{0.4}$} & {$87.3_{0.5}$} \\
        \modeladapterdeb & {$84.6_{0.2}$} & {$60.8_{0.4}$} & {$84.4_{0.0}$} & {$65.6_{0.2}$} & {$80.5_{0.7}$} & {$65.6_{0.1}^{\clubsuit}$} & {$80.1_{0.4}$} & {$82.1_{1.3}^{\clubsuit}$} \\
        \cdashlinelr{1-9}
        \modeldiff & {$84.6_{0.1}$} & {$68.9_{0.2}$} & {$84.6_{0.1}$} & {$68.9_{0.2}$} & {$81.3_{0.3}$} & {$93.2_{0.3}$} & {$81.3_{0.3}$} & {$93.2_{0.3}$} \\
        \modeldiffdeb & {$84.5_{0.1}$} & {$62.4_{0.3}$} & {$84.2_{0.4}$} & {$63.2_{1.4}$} & {$81.6_{0.4}$} & {$66.8_{3.7}$} & {$81.3_{0.2}$} & {$91.8_{0.2}$} \\
        \modelmodularseq & {$84.5_{0.1}$} & {$61.6_{0.7}$} & {$84.3_{0.1}$} & {$64.5_{0.4} $} & {$81.3_{0.2}$} & $\mathbf{66.0}_{1.2}$ & {$81.2_{0.2}$} & {$91.1_{0.9}$} \\
        \modelmodularpar & {$84.2_{0.2}$} & $\mathbf{53.7}_{1.5}^{\clubsuit}$ & {$84.5_{0.2}$} & $\mathbf{58.8}_{1.2}^{\clubsuit}$ & {$81.2_{0.6}$} & {$75.4_{3.7}$} & {$81.3_{0.7}$} & $\mathbf{85.5}_{0.4}$  \\
        \bottomrule
    \end{tabular}
  \vspace{-2mm}
  \caption{Results of the BIOS and FCDL18 datasets on \textbf{BERT-Base} with adversarial bias removal and mutual information (MI) reduction. Task performance is measured with accuracy, and bias mitigation with balanced accuracy of the probes. The protected attribute is gender for BIOS, and race-dialect for FCDL18. The results with the best bias mitigation performance (lowest values) among the models that use \emph{diff} subnetworks (lower part of the table) are shown in \textbf{bold}, and among all models with the $\clubsuit$ symbol. Subscript values indicate standard deviation.}
  \label{tab:results_other_base}
  \vspace{-6mm}
\end{table*}

\paragraph{Datasets}
We evaluate our approach on three datasets on the tasks of occupation prediction, hate speech detection, and mention prediction; involving protected attributes of gender, age, and race dialect. The first dataset is \textbf{BIOS}~ \cite{dearteaga_2019_bios} which contains short biographies used to predict a person's job, where the name and any indication of the person's gender (such as pronouns) in the biography are omitted. The BIOS dataset contains around 430K data points with 28 occupations, and two protected attribute classes (female/male). The second dataset is \textbf{FDCL18}~\cite{founta_2018_twitter} for hate speech detection, containing a set of tweets each classified as hateful, abusive, spam, or none. As discussed in \citet{xia-etal-2020-demoting}, hate speech might have a strong correlation with dialect-based racial bias. Following previous studies~\cite{sap_2019_risk,ravfogel_2020_null}, we assign race dialect labels of \emph{African American} and \emph{White American} to FDCL18 using the probabilistic model developed by \citet{blodgett_2016_demographic}, resulting in the dataset of approximately 62K data points. The third dataset is \textbf{PAN16}~\cite{rangel_2016_pan16} containing a set of tweets accompanied with the labels of gender and age of the authors. The task's objective is to predict mentions (whether another user is mentioned in a tweet). PAN16 provides approximately 200K data points with binary task classes (\emph{mention}, \emph{no mention}), as well as two gender labels and five age groups. Further details on the three datasets are provided in Appendix~\ref{sec:appendix:setting}.

\paragraph{Models and Baselines}
We conduct the experiments on the following models and baselines. \textbf{\modelfine:} finetuning all parameters of the PLM on the task without any bias mitigation objective. \textbf{\modelfinedeb:} the same model as \modelfine but with the bias mitigation objective. \textbf{\modeladapter:} learning the task with an adapter network while the rest of the PLM's parameters are kept frozen. \textbf{\modeladapterdeb:} the same model as \modeladapter but the adapter is trained on both task and bias mitigation objectives. \textbf{\modeldiff:} using a \emph{diff} network to learn the task while PLM parameters remain unchanged. \textbf{\modeldiffdeb:} same as \modeldiff but the \emph{diff} network is trained on both task and bias mitigation objectives. \textbf{\modelmodularseq:} our introduced post-hoc approach where the debiasing modules are learned after training the model. We use \modelfine as the base model for \modelmodularseq, whose parameters are kept frozen during post-hoc training. \textbf{\modelmodularpar:} our introduced parallel approach where the debiasing modules are learned together with the base model finetuned on the task. Additionally, to provide a comprehensive view on bias mitigation methods, we evaluate the datasets on the \makebox{\textbf{\modelinlp}}~\cite{ravfogel_2020_null} approach using the implementation and suggested hyperparameter setting. The evaluation results of the INLP method are separately reported in Table~\ref{tbl:results:inlp} in Appendix~\ref{sec:appendix:results}. As the PLM encoder for all models, we using two versions of BERT~\cite{devlin_2019_bert} with different sizes, namely BERT-Mini~\cite{turc_2019_bert_small} and BERT-Base. This provides us a more comprehensive picture regarding the effect of encoder size and number of involved parameters on the bias mitigation methods. 


All debiasing models are separately trained according to the adversarial bias mitigation and mutual information reduction methods. We particularly opt for a non-linear adversarial head with two fully connected layers and the $\text{tanh}$ activation. Furthermore, to improve the capacity of adversarial learning, we initialize 5 instances of $h_{\rho_i}$ and calculate the average of the loss for the backward pass. In MI reduction, in the case of protected attributes with more than two classes, we turn the multi-classes setting to multiple one-versus-rest splits. For the models with \emph{diff} subnetworks, we conduct preliminary experiments to find proper thresholds for magnitude pruning that improves sparsity as much as possible without sacrificing performance. For BERT-Base and BERT-Mini, we set the minimum sparsity threshold to 99\% and 95\% (maximum size of 1\% and 5\%), respectively. We note that these ratios are the lower bounds, and the $L_0$ regularization may by itself reach a higher sparsity. The complete details of our hyperparameters setting and training procedure are explained in Appendix~\ref{sec:appendix:setting}. Our code and trained resources are available in \textbf{\url{https://github.com/CPJKU/ModularizedDebiasing}}.

\begin{table*}[hbt!]
\centering
    \begin{tabular}{l ccc ccc}
        \toprule
        \multirow{2}{*}{Model} &
        \multicolumn{3}{c}{Adversarial Bias Mitigation} & \multicolumn{3}{c}{MI Reduction} \\
        & Task$\uparrow$ & $\text{Probe}_{\verb|G|}$$\downarrow$ &  $\text{Probe}_{\verb|A|}$$\downarrow$ & Task$\uparrow$ & $\text{Probe}_{\verb|G|}$$\downarrow$ &  $\text{Probe}_{\verb|A|}$$\downarrow$ \\
        \midrule
        
        \modelfine & {$93.5_{0.2}$} & {$70.4_{2.2}$} & {$53.6_{3.2}$} & {$93.5_{0.2}$} & {$70.4_{2.2}$} & {$53.6_{3.2}$} \\
        \modeladapter & {$87.3_{0.1}$} & {$67.8_{0.6}$} & {$37.1_{1.7}$} & {$87.3_{0.1}$} & {$67.8_{0.6}$} & {$37.1_{1.7}$} \\
        $\modelfinedeb_{\verb|G|}$ & {$93.7_{0.0}$} & {$52.3_{0.5}$} & {$34.8_{1.1}$} &{$93.7_{0.1}$} &{$62.2_{0.1}$} &{$41.9_{1.4}$} \\
        $\modelfinedeb_{\verb|A|}$ & {$92.9_{0.2}$} & {$52.3_{0.3}$} & {$31.4_{2.6}$} &{$93.7_{0.1}$} & $61.5_{0.7}$ & $41.2_{2.1}$ \\
        $\modelfinedeb_{\verb|G&A|}$ & {$93.0_{0.2}$} & {$51.7_{1.4}$} & {$33.9_{1.8}$} &$93.5_{0.1}$ &$61.4_{0.7}$ &$40.8_{1.0}$ \\
        $\modeladapterdeb_{\verb|G|}$ & {$86.9_{0.1}$} & {$54.9_{0.4}$} & {$29.2_{1.7}$} & {$87.0_{0.1}$} & {$67.4_{0.3}$} & {$37.0_{2.2}$} \\
        $\modeladapterdeb_{\verb|A|}$ & {$86.1_{0.2}$} & {$58.5_{0.2}$} & {$25.6_{2.0}^{\clubsuit}$} & {$87.1_{0.3}$} & {$67.1_{0.5}$} & {$36.6_{1.2}$} \\
        $\modeladapterdeb_{\verb|G&A|}$ & {$92.2_{0.2}$} & {$51.6_{2.4}^{\clubsuit}$} & {$27.5_{1.9}$} & {$92.8_{0.1}$} & {$64.2_{0.4}$} & {$34.9_{0.2}$} \\
        \cdashlinelr{1-7}
        \modeldiff & {$93.0_{0.1}$} & {$76.1_{0.4}$} & {$62.4_{0.5}$} & {$93.0_{0.1}$} & {$76.1_{0.4}$} & {$62.4_{0.5}$} \\
        $\modeldiffdeb_{\verb|G|}$ & {$93.4_{0.1}$} & {$56.3_{0.8}$} & {$48.5_{1.2}$} &{$93.3_{0.0}$} &{$73.0_{0.7}$} & {$58.2_{0.5}$} \\
        $\modeldiffdeb_{\verb|A|}$ & {$92.9_{0.1}$} & {$64.5_{0.8}$} & {$34.1_{1.1}$} &{$93.4_{0.0}$} &{$73.1_{0.8}$} &{$57.1_{0.6}$} \\
        $\modeldiffdeb_{\verb|G&A|}$ & {$92.9_{0.2}$} & {$54.7_{2.6}$} & {$29.7_{2.0}$} &$93.6_{0.2}$  &$73.3_{0.3}$  &$57.2_{0.9}$  \\
        $\modelmodularseq_{\verb|G|}$ & {$93.7_{0.1}$} & {$57.0_{1.0}$} & {$45.4_{1.3}$} &$93.6_{0.1}$ &$66.2_{0.2}$ &$46.2_{1.4}$ \\
        $\modelmodularseq_{\verb|A|}$ & {$93.7_{0.1}$} & {$63.7_{0.9}$} & {$31.7_{2.8}$} &$93.6_{0.0}$ &$66.6_{0.2}$ &$46.5_{2.8}$ \\
        $\modelmodularseq_{\verb|G|}+\emph{ditto}_{\verb|A|}$ & {$92.3_{0.6}$} & {$57.7_{1.2}$} & {$32.0_{2.8}$} &$93.4_{0.1}$ &$66.4_{0.8}$ &$46.6_{1.8}$ \\
        $\modelmodularseq_{\verb|G&A|}$ & {$93.6_{0.1}$} & {$52.6_{2.2}$} & {$30.9_{0.5}$} &$93.6_{0.1}$  &$66.9_{0.4}$ &$47.4_{0.1}$ \\
        $\modelmodularpar_{\verb|G|}$ & {$93.5_{0.2}$} & {$53.0_{1.6}$} & {$32.2_{1.1}$} &$93.6_{0.0}$  &$59.3_{0.4}$  &$35.7_{0.4}$  \\
        $\modelmodularpar_{\verb|A|}$ & {$93.5_{0.2}$} & {$53.8_{1.9}$} & {$30.1_{1.0}$} &$93.7_{0.2}$  &$55.0_{0.2}$  &$29.9_{0.9}$  \\
        $\modelmodularpar_{\verb|G|}+\emph{ditto}_{\verb|A|}$ & {$93.5_{0.2}$} & {$52.8_{1.7}$} & {$30.2_{0.8}$} & {$93.6_{0.2}$} & {$56.1_{1.1}$} & {$30.1_{1.0}$} \\
        $\modelmodularpar_{\verb|G&A|}$ & {$93.8_{0.2}$} & {$\mathbf{52.3}_{1.4}$} & {$\mathbf{28.3}_{1.6}$} & $93.6_{0.2}$ & $\mathbf{52.7}_{1.3}^{\clubsuit}$ & $\mathbf{29.4}_{0.6}^{\clubsuit}$ \\
        \bottomrule
    \end{tabular}
  \vspace{-1mm}
  \caption{Results of the PAN16 dataset on \textbf{BERT-Base}. The subscripts \texttt{G} and \texttt{A} refer to the protected attributes gender and age, respectively. The sign \texttt{G\&A} denotes that the bias mitigation loss is the sum of the debiasing loss terms of both gender and age. The \modelmodular models in the form of ``$+\emph{ditto}$'' refer to the case where first two debiasing subnetworks with the same core model are trained separately, and then they are added to the base model at inference time. The results with the best bias mitigation performance (lowest values) among the models that use \emph{diff} subnetworks (lower part of the table) are shown in \textbf{bold}, and among all models with the $\clubsuit$ symbol. Subscript values indicate standard deviation.}
  \vspace{-5mm}
  \label{tab:results_pan16_base}
\end{table*}

\paragraph{Evaluation Metrics}
We evaluate the performance of the classifiers on the core task using the accuracy metric. We evaluate bias mitigation based on the concept of \emph{fairness through blindness}, namely by examining whether models are agnostic about the protected attributes. Concretely following previous works~\cite{elazar_2018_adv,barrett_2019_adversarial}, we report the leakage of a protected attribute in terms of the performance of a strong probe (or attacker) network. To train the probe, we freeze the model's parameters, and train a new classification head (two-layer feed-forward layer with a $\text{tanh}$ activation) to predict the protected attribute from the $\vz$ encoding vector. For each evaluation, we train an ensemble of 5 probes for 40 epochs with early stopping if validation loss does not increase over 5 epochs, and report the results of the best performing probe. We report the performance of the probe in terms of balanced/macro accuracy (average of per-class accuracy scores). Balanced accuracy has the benefit of better reflecting the performance of the methods when considering minority groups, particularly given the unbalanced distributions over protected labels in the datasets. To account for possible variabilities, we repeat every experiment five times and report the mean and standard deviation.

\section{Results and Analysis}
\label{sec:results}

\paragraph{Single-attribute Evaluation} Table~\ref{tab:results_other_base} reports the evaluation results of the BIOS and FCDL18 datasets on BERT-Base using adversarial bias removal and mutual information (MI) reduction. The results of the same experiments on BERT-Mini are shown in Table~\ref{tab:results_other_mini} in Appendix~\ref{sec:appendix:results}.

Starting from task accuracy, the models using subnetworks (variations of \modeldiff and \modelmodular shown at the lower part of the table) consistently perform the same as the fully finetuned models on both datasets and debiasing methods. We observe a slight decrease in performance for the adapter-based models on FCDL18. 

Looking at the leakage probing performance of bias attributes, \modelmodular models show better performance (lower values) among the subnetwork-based models on both datasets, and overall on BIOS.\footnote{Particularly on FCDL18, we observe high variations, which requires us to more cautiously interpret results. We assume that this is due to the small size of this dataset especially in the learning regimes with many parameters on BERT-Base, as this effect is less pronounced on BERT-Mini (Table~\ref{tab:results_other_mini} in Appendix~\ref{sec:appendix:results}).}
 In particular, \modelmodular models outperform the directly comparable baselines \modelfinedeb and \modeldiffdeb on all configurations, indicating the benefits of learning separate debiasing modules on bias mitigation performance. This indeed comes with the core advantages of \modelmodular models in proving modularized and on-demand bias mitigation.

 The results also show that \modelmodularseq, while (as expected) slightly weaker than \modelmodularpar, provides competitive bias mitigation performance (\ie on par with \modeldiffdeb). Finally, comparing between debiasing optimizations, MI reduction shows consistently worse bias mitigation performance in comparison with adversarial training, particularly on FCDL18 where the protected attribute has more than two labels. 

\paragraph{Two-attribute Evaluation} The results on PAN16 using BERT-Base are reported in Table~\ref{tab:results_pan16_base}, and the same experiments on BERT-Mini in Table~\ref{tab:results_pan16_mini} in Appendix~\ref{sec:appendix:results}. In this experiment, every debiasing model is trained based on either gender, age, or simultaneously on both gender and age, shown with the subscripts \verb|G|, \verb|A|, \verb|G&A|, respectively. An additional evaluation is indicated with ${\modelmodularstar}_{\verb|G|}+\emph{ditto}_{\verb|A|}$, which refers to adding the two separately trained gender and age debiasing subnetworks to the core model at inference time. In fact, for the experiments of each \modelmodular model indicated with ``$+\emph{ditto}$'', \verb|G|, and \verb|A|, we train only one model with gender and age subnetworks, and then add the respective subnetwork(s) to the core model.

Looking at task performance results, similar to the previous datasets the models perform on par with fully finetuning, except the adapter-based models which in this case significantly underperform in both optimization methods. Regarding bias mitigation performance, we observe similar patterns to the ones discussed on the other datasets: \modelmodular models particularly \modelmodularpar show the least attribute leakage among the subnetwork-based models consistently, and also over all models with MI reduction. This reaffirms the benefits of modularizing debiasing of the attributes separately from the task. The results also indicate the existence of a correlation between the gender and age attributes in this dataset, such that debiasing each attribute also results in a decrease in leakage of the other attribute. The overall best results are achieved on the models that simultaneously optimize on both attributes (\verb|G&A|). 

Finally, let us have a closer look at the results of applying the two independently trained subnetworks. On both training regimes, we observe on par debiasing performance between ${\modelmodularstar}_{\verb|G|}+\emph{ditto}_{\verb|A|}$ and the corresponding results with one subnetwork, namely the ones of gender and age debiasing in ${\modelmodularstar}_{\verb|G|}$ and ${\modelmodularstar}_{\verb|A|}$, respectively. These results indicate the viability of our approach to effectively merge subnetworks at inference time.


\begin{table}[t]
\small
\centering
\begin{tabular}{l cccc}
\toprule
& \multicolumn{2}{c}{PAN16} & BIOS & FCDL18 \\
& Gender & Age & Gender & Dialect \\
\midrule
Overall & 0.01\% & 1.00\% & 0.27\% & 0.18\% \\
\cdashlinelr{1-5}
Layer 12 & 0.04\% & 7.00\% & 0.26\% & 0.92\% \\
Layer 11 & 0.02\% & 3.59\% & 0.30\% & 0.52\% \\
Layer 10 & 0.01\% & 1.76\% & 0.33\% & 0.30\% \\
Layer 9 & 0.00\% & 0.49\% & 0.20\% & 0.18\% \\
Layer 8 & 0.00\% & 0.38\% & 0.22\% & 0.11\% \\
Layer 7 & 0.01\% & 0.21\% & 0.30\% & 0.14\% \\
Layer 6 & 0.00\% & 0.15\% & 0.28\% & 0.13\% \\
Layer 5 & 0.01\% & 0.03\% & 0.28\% & 0.07\% \\
Layer 4 & 0.01\% & 0.25\% & 0.43\% & 0.08\% \\
Layer 3 & 0.01\% & 0.25\% & 0.56\% & 0.08\% \\
Layer 2 & 0.01\% & 0.38\% & 0.41\% & 0.07\% \\
Layer 1 & 0.01\% & 0.21\% & 0.44\% & 0.06\% \\
Embeddings & 0.00\% & 0.04\% & 0.05\% & 0.03\% \\
\bottomrule
\end{tabular}
\vspace{-2mm}
\caption{The percentage of non-masked parameters in the debiasing subnetworks of \modelmodularpar.}
\label{tbl:results:sparsity_rate}
\vspace{-5mm}
\end{table}

\paragraph{Subnetworks analysis} We further investigate the achieved sparsity rate of the subnetworks, keeping in mind that the maximum capacity of the debiasing subnetworks on BERT-Base is set to 1\%. The size of a subnetwork indicates the amount of information or in fact modifications needed to be applied, in order to debias a protected attribute. Table~\ref{tbl:results:sparsity_rate} reports the percentage of the number of non-masked parameters in every layer, and also overall, in the subnetworks of the \modelmodularpar models regarding the protected attributes. The results show interesting patterns in respect to various protected attributes: the gender attribute on both PAN16 and BIOS dataset require much smaller subnetworks, such that the subnetwork on PAN16 is only 0.01\% of the size of the core model. The age attribute appears to be a more complex topic in the underlying PLM, as it fully uses the 1\% maximum capacity. Looking across the layers, the results show that debiasing the gender attribute is mostly handled in the lower transformer layers (particularly on BIOS), while debiasing age and dialect attributes mostly happens at the higher layers. 

In Appendix~\ref{sec:appendix:nonzero}, we further discuss this topic for all models on the level of individual weight matrices. Moreover, in   Appendix~\ref{sec:appendix:homogeneity} we investigate to what extent the subnetworks of a model across several runs affect on the same set of parameters.

\section{Conclusion}
We propose \modelmodular, a novel bias mitigation approach which enables integration of an arbitrary subset of the debiasing modules at inference time. Our method encapsulates the functionality of bias mitigation in respect to a protected attribute into a separate magnitude-difference subnetwork, which can then be applied to the core model on-demand. Our experiments on three classification tasks show that 
\modelmodular improves bias mitigation achieved by separating the debiasing from the task network, and effectively mitigates the bias of two (and possibly more) attributes when their respective subnetworks are simultaneously utilized. 

\section{Limitations}
An important limitation of our work concerns the definition of the protected attributes in the datasets used for evaluation. In particular, gender in BIOS and PAN16 is limited to the binary female/male, lacking an inclusive and nuanced definition of gender. Similarly in FDCL18, we consider only two dialects of \emph{African American} and \emph{White American}, while clearly this definition is limited and non-inclusive. Furthermore as in previous work~\cite{sap_2019_risk,ravfogel_2020_null, zhang_2021_disentangling}, the labels of this protected attribute are assigned through a probabilistic model, and hence the dataset might not represent the nuances and traits of the real-world.

The second limitation regards reaching strong conclusions on the generalizability of the multi-attribute setting for \modelmodular over any possible number of protected attributes or subset of them. Our multi-attribute experiments are conducted on one dataset with two attributes of gender and age, particularly due to the lack of available suitable datasets. Hence, Further studies (as well as more suitable datasets) are required for achieving a more comprehensive picture on the topic.


Finally, we should also highlight two general limitations, shared with the other related studies in the area of model bias mitigation. First, we should consider that the aim of representation disentanglement optimizations is to reduce the existing \emph{correlations} in the model with the protected attributes based on the \emph{observed data}. These data-oriented approaches might lack effective generalization, particularly when the model is evaluated in other domains or out-of-distribution data. Second, our bias mitigation evaluation is grounded in the notion of \emph{fairness through blindness}, and the debiasing optimization methods are designed to support this form of fairness. The effects of our method on other possible definitions of fairness are therefore left for future work. 

\section{Acknowledgment}
This work received financial support by the Austrian Science Fund (FWF): P33526 and DFH-23; and by the State of Upper Austria and the Federal Ministry of Education, Science, and Research, through grants LIT-2020-9-SEE-113 and LIT-2021-YOU-215.

\bibliography{references}

\begin{thebibliography}{59}
\expandafter\ifx\csname natexlab\endcsname\relax\def\natexlab#1{#1}\fi

\bibitem[{Barrett et~al.(2019)Barrett, Kementchedjhieva, Elazar, Elliott, and
  S{\o}gaard}]{barrett_2019_adversarial}
Maria Barrett, Yova Kementchedjhieva, Yanai Elazar, Desmond Elliott, and Anders
  S{\o}gaard. 2019.
\newblock Adversarial removal of demographic attributes revisited.
\newblock In \emph{Proceedings of the Conference on Empirical Methods in
  Natural Language Processing and the 9th International Joint Conference on
  Natural Language Processing (EMNLP-IJCNLP)}, pages 6331--6336.

\bibitem[{Ben{-}Zaken et~al.(2021)Ben{-}Zaken, Ravfogel, and
  Goldberg}]{zaken_2021_bitfit}
Elad Ben{-}Zaken, Shauli Ravfogel, and Yoav Goldberg. 2021.
\newblock Bitfit: Simple parameter-efficient fine-tuning for transformer-based
  masked language-models.
\newblock \emph{ArXiv}, abs/2106.10199.

\bibitem[{Biega et~al.(2018)Biega, Gummadi, and Weikum}]{biega2018equity}
Asia~J Biega, Krishna~P Gummadi, and Gerhard Weikum. 2018.
\newblock Equity of attention: Amortizing individual fairness in rankings.
\newblock In \emph{The 41st international ACM SIGIR Conference on Research \&
  Development in Information Retrieval}, pages 405--414.

\bibitem[{Blodgett et~al.(2020)Blodgett, Barocas, Daum{\'e}~III, and
  Wallach}]{blodgett2020language}
Su~Lin Blodgett, Solon Barocas, Hal Daum{\'e}~III, and Hanna Wallach. 2020.
\newblock Language (technology) is power: A critical survey of “bias” in
  nlp.
\newblock In \emph{Proceedings of the 58th Annual Meeting of the Association
  for Computational Linguistics}, pages 5454--5476.

\bibitem[{Blodgett et~al.(2016)Blodgett, Green, and
  O{'}Connor}]{blodgett_2016_demographic}
Su~Lin Blodgett, Lisa Green, and Brendan O{'}Connor. 2016.
\newblock \href {https://doi.org/10.18653/v1/D16-1120} {Demographic dialectal
  variation in social media: A case study of {A}frican-{A}merican {E}nglish}.
\newblock In \emph{Proceedings of the 2016 Conference on Empirical Methods in
  Natural Language Processing}, pages 1119--1130, Austin, Texas. Association
  for Computational Linguistics.

\bibitem[{Bolukbasi et~al.(2016)Bolukbasi, Chang, Zou, Saligrama, and
  Kalai}]{bolukbasi_2016_bias_embedding}
Tolga Bolukbasi, Kai-Wei Chang, James~Y Zou, Venkatesh Saligrama, and Adam~T
  Kalai. 2016.
\newblock \href
  {https://proceedings.neurips.cc/paper/2016/file/a486cd07e4ac3d270571622f4f316ec5-Paper.pdf}
  {Man is to computer programmer as woman is to homemaker? debiasing word
  embeddings}.
\newblock In \emph{Advances in Neural Information Processing Systems},
  volume~29. Curran Associates, Inc.

\bibitem[{Chen et~al.(2020)Chen, Frankle, Chang, Liu, Zhang, Wang, and
  Carbin}]{lottery_ticket_BERT_pruning_Chen2020}
Tianlong Chen, Jonathan Frankle, Shiyu Chang, Sijia Liu, Yang Zhang, Zhangyang
  Wang, and Michael Carbin. 2020.
\newblock The lottery ticket hypothesis for pre-trained {BERT} networks.
\newblock In \emph{Proceedings of NeurIPS}.

\bibitem[{Cheng et~al.(2020)Cheng, Min, Shen, Malon, Zhang, Li, and
  Carin}]{cheng_2020_improving}
Pengyu Cheng, Martin~Renqiang Min, Dinghan Shen, Christopher Malon, Yizhe
  Zhang, Yitong Li, and Lawrence Carin. 2020.
\newblock \href {https://doi.org/10.18653/v1/2020.acl-main.673} {Improving
  disentangled text representation learning with information-theoretic
  guidance}.
\newblock In \emph{Proceedings of the 58th Annual Meeting of the Association
  for Computational Linguistics}, pages 7530--7541, Online. Association for
  Computational Linguistics.

\bibitem[{Colombo et~al.(2021)Colombo, Piantanida, and
  Clavel}]{colombo_2021_novel}
Pierre Colombo, Pablo Piantanida, and Chlo{\'e} Clavel. 2021.
\newblock A novel estimator of mutual information for learning to disentangle
  textual representations.
\newblock In \emph{Proceedings of the 59th Annual Meeting of the Association
  for Computational Linguistics and the 11th International Joint Conference on
  Natural Language Processing (Volume 1: Long Papers)}, pages 6539--6550.

\bibitem[{De-Arteaga et~al.(2019)De-Arteaga, Romanov, Wallach, Chayes, Borgs,
  Chouldechova, Geyik, Kenthapadi, and Kalai}]{dearteaga_2019_bios}
Maria De-Arteaga, Alexey Romanov, Hanna Wallach, Jennifer Chayes, Christian
  Borgs, Alexandra Chouldechova, Sahin Geyik, Krishnaram Kenthapadi, and
  Adam~Tauman Kalai. 2019.
\newblock \href {https://doi.org/10.1145/3287560.3287572} {Bias in bios: A case
  study of semantic representation bias in a high-stakes setting}.
\newblock page 120–128, New York, NY, USA. Association for Computing
  Machinery.

\bibitem[{Devlin et~al.(2019)Devlin, Chang, Lee, and
  Toutanova}]{devlin_2019_bert}
Jacob Devlin, Ming-Wei Chang, Kenton Lee, and Kristina Toutanova. 2019.
\newblock \href {https://doi.org/10.18653/v1/N19-1423} {Bert: Pre-training of
  deep bidirectional transformers for language understanding}.
\newblock In \emph{Conference of the North American Chapter of the Association
  for Computational Linguistics: Human Language Technologies}, volume~1, pages
  4171--4186. Association for Computational Linguistics.

\bibitem[{Elazar and Goldberg(2018)}]{elazar_2018_adv}
Yanai Elazar and Yoav Goldberg. 2018.
\newblock \href {https://doi.org/10.18653/v1/D18-1002} {Adversarial removal of
  demographic attributes from text data}.
\newblock In \emph{Proceedings of the 2018 Conference on Empirical Methods in
  Natural Language Processing}, pages 11--21. Association for Computational
  Linguistics.

\bibitem[{Founta et~al.(2018)Founta, Djouvas, Chatzakou, Leontiadis, Blackburn,
  Stringhini, Vakali, Sirivianos, and Kourtellis}]{founta_2018_twitter}
Antigoni~Maria Founta, Constantinos Djouvas, Despoina Chatzakou, Ilias
  Leontiadis, Jeremy Blackburn, Gianluca Stringhini, Athena Vakali, Michael
  Sirivianos, and Nicolas Kourtellis. 2018.
\newblock Large scale crowdsourcing and characterization of twitter abusive
  behavior.
\newblock In \emph{Twelfth International AAAI Conference on Web and Social
  Media}.

\bibitem[{Frankle and Carbin(2019)}]{lottery_ticket_original_Frankle2019}
Jonathan Frankle and Michael Carbin. 2019.
\newblock The lottery ticket hypothesis: Finding sparse, trainable neural
  networks.
\newblock In \emph{Proceedings of International Conference on Learning
  Representations, {ICLR}}.

\bibitem[{Ganh\"{o}r et~al.(2022)Ganh\"{o}r, Penz, Rekabsaz, Lesota, and
  Schedl}]{ganhoer2022mitigating}
Christian Ganh\"{o}r, David Penz, Navid Rekabsaz, Oleg Lesota, and Markus
  Schedl. 2022.
\newblock \href {https://doi.org/10.1145/3477495.3531820} {Unlearning protected
  user attributes in recommendations with adversarial training}.
\newblock In \emph{Proceedings of the 45th International ACM SIGIR Conference
  on Research and Development in Information Retrieval}, SIGIR '22, page
  2142–2147, New York, NY, USA. Association for Computing Machinery.

\bibitem[{Ganin and Lempitsky(2015)}]{ganin_2015_domainadaptation}
Yaroslav Ganin and Victor Lempitsky. 2015.
\newblock Unsupervised domain adaptation by backpropagation.
\newblock In \emph{International conference on machine learning}, pages
  1180--1189. Proceedings of Machine Learning Research.

\bibitem[{Ganin et~al.(2016)Ganin, Ustinova, Ajakan, Germain, Larochelle,
  Laviolette, Marchand, and Lempitsky}]{ganin_2016_domain_adv}
Yaroslav Ganin, Evgeniya Ustinova, Hana Ajakan, Pascal Germain, Hugo
  Larochelle, François Laviolette, Mario Marchand, and Victor Lempitsky. 2016.
\newblock Domain-adversarial training of neural networks.
\newblock In \emph{Journal of Machine Learning Research}, volume~17, pages
  1--35.

\bibitem[{Gretton et~al.(2012)Gretton, Borgwardt, Rasch, Sch{\"o}lkopf, and
  Smola}]{gretton:mmd}
Arthur Gretton, Karsten~M Borgwardt, Malte~J Rasch, Bernhard Sch{\"o}lkopf, and
  Alexander Smola. 2012.
\newblock A kernel two-sample test.
\newblock \emph{The Journal of Machine Learning Research}, 13(1):723--773.

\bibitem[{Guo et~al.(2021)Guo, Rush, and Kim}]{guo_2021_diffpruning}
Demi Guo, Alexander Rush, and Yoon Kim. 2021.
\newblock Parameter-efficient transfer learning with diff pruning.
\newblock In \emph{Proceedings of the 59th Annual Meeting of the Association
  for Computational Linguistics}, pages 4884--4896.

\bibitem[{Guo et~al.(2022)Guo, Yang, and Abbasi}]{guo_2022_auto}
Yue Guo, Yi~Yang, and Ahmed Abbasi. 2022.
\newblock Auto-debias: Debiasing masked language models with automated biased
  prompts.
\newblock In \emph{Proceedings of the 60th Annual Meeting of the Association
  for Computational Linguistics (Volume 1: Long Papers)}, pages 1012--1023.

\bibitem[{Han et~al.(2021{\natexlab{a}})Han, Pang, and Wu}]{han2021robust}
Wenjuan Han, Bo~Pang, and Ying~Nian Wu. 2021{\natexlab{a}}.
\newblock Robust transfer learning with pretrained language models through
  adapters.
\newblock In \emph{Proceedings of the 59th Annual Meeting of the Association
  for Computational Linguistics and the 11th International Joint Conference on
  Natural Language Processing (Volume 2: Short Papers)}, pages 854--861.

\bibitem[{Han et~al.(2021{\natexlab{b}})Han, Baldwin, and
  Cohn}]{han_2021_diverse}
Xudong Han, Timothy Baldwin, and Trevor Cohn. 2021{\natexlab{b}}.
\newblock \href {https://doi.org/10.18653/v1/2021.eacl-main.239} {Diverse
  adversaries for mitigating bias in training}.
\newblock In \emph{Proceedings of the 16th Conference of the European Chapter
  of the Association for Computational Linguistics: Main Volume}, pages
  2760--2765, Online. Association for Computational Linguistics.

\bibitem[{Houlsby et~al.(2019)Houlsby, Giurgiu, Jastrzebski, Morrone,
  De~Laroussilhe, Gesmundo, Attariyan, and
  Gelly}]{houlsby_2019_param_efficient}
Neil Houlsby, Andrei Giurgiu, Stanislaw Jastrzebski, Bruna Morrone, Quentin
  De~Laroussilhe, Andrea Gesmundo, Mona Attariyan, and Sylvain Gelly. 2019.
\newblock \href {https://proceedings.mlr.press/v97/houlsby19a.html}
  {Parameter-efficient transfer learning for {NLP}}.
\newblock In \emph{International Conference on Machine Learning}, volume~97,
  pages 2790--2799. Proceedings of Machine Learning Research.

\bibitem[{Hu et~al.(2021)Hu, Shen, Wallis, Allen-Zhu, Li, Wang, and
  Chen}]{hu_2021_lora}
Edward Hu, Yelong Shen, Phil Wallis, Zeyuan Allen-Zhu, Yuanzhi Li, Lu~Wang, and
  Weizhu Chen. 2021.
\newblock \href {http://arxiv.org/abs/2106.09685} {Lora: Low-rank adaptation of
  large language models}.

\bibitem[{Jaszczur et~al.(2021)Jaszczur, Chowdhery, Mohiuddin, Kaiser,
  Gajewski, Michalewski, and Kanerva}]{jaszczur2021sparse}
Sebastian Jaszczur, Aakanksha Chowdhery, Afroz Mohiuddin, {\L}ukasz Kaiser,
  Wojciech Gajewski, Henryk Michalewski, and Jonni Kanerva. 2021.
\newblock Sparse is enough in scaling transformers.
\newblock \emph{Advances in Neural Information Processing Systems}, 34.

\bibitem[{Kaneko and Bollegala(2021)}]{kaneko_2021_debiasing}
Masahiro Kaneko and Danushka Bollegala. 2021.
\newblock Debiasing pre-trained contextualised embeddings.
\newblock In \emph{Proceedings of the 16th Conference of the European Chapter
  of the Association for Computational Linguistics: Main Volume}, pages
  1256--1266.

\bibitem[{Krieg et~al.(2023)Krieg, Parada-Cabaleiro, Medicus, Lesota, Schedl,
  and Rekabsaz}]{krieg2022grep}
Klara Krieg, Emilia Parada-Cabaleiro, Gertraud Medicus, Oleg Lesota, Markus
  Schedl, and Navid Rekabsaz. 2023.
\newblock Grep-biasir: A dataset for investigating gender representation-bias
  in information retrieval results.
\newblock In \emph{Proceeding of the ACM SIGIR Conference On Human Information
  Interaction And Retrieval (CHIIR)}.

\bibitem[{Lagunas et~al.(2021)Lagunas, Charlaix, Sanh, and
  Rush}]{lagunas2021block}
Fran{\c{c}}ois Lagunas, Ella Charlaix, Victor Sanh, and Alexander~M Rush. 2021.
\newblock Block pruning for faster transformers.
\newblock In \emph{Proceedings of the 2021 Conference on Empirical Methods in
  Natural Language Processing}, pages 10619--10629.

\bibitem[{Lauscher et~al.(2021)Lauscher, Lueken, and
  Glava{\v{s}}}]{lauscher_2021_sustainable}
Anne Lauscher, Tobias Lueken, and Goran Glava{\v{s}}. 2021.
\newblock \href {https://doi.org/10.18653/v1/2021.findings-emnlp.411}
  {Sustainable modular debiasing of language models}.
\newblock In \emph{Findings of the Association for Computational Linguistics:
  EMNLP 2021}, pages 4782--4797, Punta Cana, Dominican Republic. Association
  for Computational Linguistics.

\bibitem[{Louizos et~al.(2018)Louizos, Welling, and Kingma}]{louizos_2018_l0}
Christos Louizos, Max Welling, and Diederik~P. Kingma. 2018.
\newblock Learning sparse neural networks through l0 regularization.
\newblock In \emph{International Conference on Learning Representations}.

\bibitem[{Madras et~al.(2018)Madras, Creager, Pitassi, and
  Zemel}]{madras_2018_learning}
David Madras, Elliot Creager, Toniann Pitassi, and Richard Zemel. 2018.
\newblock Learning adversarially fair and transferable representations.
\newblock In \emph{Proceedings of the International Conference on Machine
  Learning}, pages 3384--3393. PMLR.

\bibitem[{Mahabadi et~al.(2021)Mahabadi, Henderson, and
  Ruder}]{DBLP:conf/nips/MahabadiHR21}
Rabeeh~Karimi Mahabadi, James Henderson, and Sebastian Ruder. 2021.
\newblock \href
  {https://proceedings.neurips.cc/paper/2021/hash/081be9fdff07f3bc808f935906ef70c0-Abstract.html}
  {Compacter: Efficient low-rank hypercomplex adapter layers}.
\newblock In \emph{Advances in Neural Information Processing Systems 34: Annual
  Conference on Neural Information Processing Systems 2021, NeurIPS 2021,
  December 6-14, 2021, virtual}, pages 1022--1035.

\bibitem[{Meissner et~al.(2022)Meissner, Sugawara, and
  Aizawa}]{meissner_2022_debias}
Johannes~Mario Meissner, Saku Sugawara, and Akiko Aizawa. 2022.
\newblock Debiasing masks: A new framework for shortcut mitigation in {NLU}.
\newblock In \emph{Proceeding of the 2022 Conference on Empirical Methods in
  Natural Language Processing (EMNLP)}.

\bibitem[{Pardo et~al.(2016)Pardo, Rosso, Verhoeven, Daelemans, Potthast, and
  Stein}]{Pardo2016OverviewOT}
Francisco Manuel~Rangel Pardo, Paolo Rosso, Ben Verhoeven, Walter Daelemans,
  Martin Potthast, and Benno Stein. 2016.
\newblock Overview of the 4th author profiling task at pan 2016: Cross-genre
  evaluations.
\newblock In \emph{CLEF}.

\bibitem[{Pfeiffer et~al.(2021)Pfeiffer, Kamath, R{\"u}ckl{\'e}, Cho, and
  Gurevych}]{pfeiffer2021adapterfusion}
Jonas Pfeiffer, Aishwarya Kamath, Andreas R{\"u}ckl{\'e}, Kyunghyun Cho, and
  Iryna Gurevych. 2021.
\newblock Adapterfusion: Non-destructive task composition for transfer
  learning.
\newblock In \emph{Proceedings of the 16th Conference of the European Chapter
  of the Association for Computational Linguistics: Main Volume}, pages
  487--503.

\bibitem[{Rangel et~al.(2016)Rangel, Rosso, Verhoeven, Daelemans, Potthast, and
  Stein}]{rangel_2016_pan16}
Francisco Rangel, Paolo Rosso, Ben Verhoeven, Walter Daelemans, Martin
  Potthast, and Benno Stein. 2016.
\newblock \href {https://doi.org/10.5281/zenodo.3745963} {Pan16 author
  profiling}.
\newblock Zenodo.

\bibitem[{Ravfogel et~al.(2020)Ravfogel, Elazar, Gonen, Twiton, and
  Goldberg}]{ravfogel_2020_null}
Shauli Ravfogel, Yanai Elazar, Hila Gonen, Michael Twiton, and Yoav Goldberg.
  2020.
\newblock Null it out: Guarding protected attributes by iterative nullspace
  projection.
\newblock In \emph{Proceedings of the 58th Annual Meeting of the Association
  for Computational Linguistics}, pages 7237--7256.

\bibitem[{Rebuffi et~al.(2017)Rebuffi, Bilen, and
  Vedaldi}]{rebuffi2017learning}
Sylvestre-Alvise Rebuffi, Hakan Bilen, and Andrea Vedaldi. 2017.
\newblock Learning multiple visual domains with residual adapters.
\newblock In \emph{Advances in Neural Information Processing Systems
  (NeurIPS)}, volume~30.

\bibitem[{Rekabsaz et~al.(2021)Rekabsaz, Kopeinik, and
  Schedl}]{rekabsaz_2021_societal}
Navid Rekabsaz, Simone Kopeinik, and Markus Schedl. 2021.
\newblock Societal biases in retrieved contents: Measurement framework and
  adversarial mitigation of {BERT} rankers.
\newblock In \emph{Proceedings of the 44th International ACM SIGIR Conference
  on Research and Development in Information Retrieval}, pages 306--316.

\bibitem[{Rekabsaz and Schedl(2020)}]{rekabsaz_2020_neural}
Navid Rekabsaz and Markus Schedl. 2020.
\newblock Do neural ranking models intensify gender bias?
\newblock In \emph{Proceedings of the 43rd International ACM SIGIR Conference
  on Research and Development in Information Retrieval}, pages 2065--2068.

\bibitem[{R{\"u}ckl{\'e} et~al.(2021)R{\"u}ckl{\'e}, Geigle, Glockner, Beck,
  Pfeiffer, Reimers, and Gurevych}]{ruckle2021adapterdrop}
Andreas R{\"u}ckl{\'e}, Gregor Geigle, Max Glockner, Tilman Beck, Jonas
  Pfeiffer, Nils Reimers, and Iryna Gurevych. 2021.
\newblock {AdapterDrop}: On the efficiency of adapters in transformers.
\newblock In \emph{Proceedings of the 2021 Conference on Empirical Methods in
  Natural Language Processing}, pages 7930--7946.

\bibitem[{Sap et~al.(2019)Sap, Card, Gabriel, Choi, and Smith}]{sap_2019_risk}
Maarten Sap, Dallas Card, Saadia Gabriel, Yejin Choi, and Noah~A. Smith. 2019.
\newblock \href {https://doi.org/10.18653/v1/P19-1163} {The risk of racial bias
  in hate speech detection}.
\newblock In \emph{Proceedings of the 57th Annual Meeting of the Association
  for Computational Linguistics}, pages 1668--1678, Florence, Italy.
  Association for Computational Linguistics.

\bibitem[{Schick et~al.(2021)Schick, Udupa, and Sch{\"u}tze}]{schick_2021_self}
Timo Schick, Sahana Udupa, and Hinrich Sch{\"u}tze. 2021.
\newblock Self-diagnosis and self-debiasing: A proposal for reducing
  corpus-based bias in {NLP}.
\newblock \emph{Transactions of the Association for Computational Linguistics},
  9:1408--1424.

\bibitem[{Sheng et~al.(2019)Sheng, Chang, Natarajan, and Peng}]{sheng2019woman}
Emily Sheng, Kai-Wei Chang, Prem Natarajan, and Nanyun Peng. 2019.
\newblock The woman worked as a babysitter: On biases in language generation.
\newblock In \emph{Proceedings of the 2019 Conference on Empirical Methods in
  Natural Language Processing and the 9th International Joint Conference on
  Natural Language Processing (EMNLP-IJCNLP)}, pages 3398--3403.

\bibitem[{Stanovsky et~al.(2019)Stanovsky, Smith, and
  Zettlemoyer}]{stanovsky2019evaluating}
Gabriel Stanovsky, Noah~A Smith, and Luke Zettlemoyer. 2019.
\newblock Evaluating gender bias in machine translation.
\newblock In \emph{Proceedings of the 57th Annual Meeting of the Association
  for Computational Linguistics}, pages 1679--1684.

\bibitem[{Stickland and Murray(2019)}]{pmlr-v97-stickland19a}
Asa~Cooper Stickland and Iain Murray. 2019.
\newblock \href {https://proceedings.mlr.press/v97/stickland19a.html} {{BERT}
  and {PAL}s: Projected attention layers for efficient adaptation in multi-task
  learning}.
\newblock In \emph{Proceedings of the 36th International Conference on Machine
  Learning}, volume~97 of \emph{Proceedings of Machine Learning Research},
  pages 5986--5995. PMLR.

\bibitem[{Sung et~al.(2021)Sung, Nair, and Raffel}]{sung_2021_fisher}
Yi-Lin Sung, Varun Nair, and Colin Raffel. 2021.
\newblock Training neural networks with fixed sparse masks.
\newblock \emph{ArXiv}, abs/2111.09839.

\bibitem[{Turc et~al.(2019)Turc, Chang, Lee, and
  Toutanova}]{turc_2019_bert_small}
Iulia Turc, Ming-Wei Chang, Kenton Lee, and Kristina Toutanova. 2019.
\newblock Well-read students learn better: On the importance of pre-training
  compact models.
\newblock \emph{arXiv preprint arXiv:1908.08962v2}.

\bibitem[{Tzeng et~al.(2014)Tzeng, Hoffman, Zhang, Saenko, and
  Darrell}]{tzeng2014deep}
Eric Tzeng, Judy Hoffman, Ning Zhang, Kate Saenko, and Trevor Darrell. 2014.
\newblock Deep domain confusion: Maximizing for domain invariance.
\newblock \emph{arXiv preprint arXiv:1412.3474}.

\bibitem[{Wang et~al.(2021)Wang, Yan, He, Wu, and Xu}]{wang_2021_dynamically}
Liwen Wang, Yuanmeng Yan, Keqing He, Yanan Wu, and Weiran Xu. 2021.
\newblock Dynamically disentangling social bias from task-oriented
  representations with adversarial attack.
\newblock In \emph{Proceedings of the 2021 Conference of the North American
  Chapter of the Association for Computational Linguistics: Human Language
  Technologies}, pages 3740--3750.

\bibitem[{Wortsman et~al.(2020)Wortsman, Ramanujan, Liu, Kembhavi, Rastegari,
  Yosinski, and Farhadi}]{wortsman2020supermasks}
Mitchell Wortsman, Vivek Ramanujan, Rosanne Liu, Aniruddha Kembhavi, Mohammad
  Rastegari, Jason Yosinski, and Ali Farhadi. 2020.
\newblock Supermasks in superposition.
\newblock \emph{Advances in Neural Information Processing Systems}, 33.

\bibitem[{Xia et~al.(2020)Xia, Field, and Tsvetkov}]{xia-etal-2020-demoting}
Mengzhou Xia, Anjalie Field, and Yulia Tsvetkov. 2020.
\newblock \href {https://doi.org/10.18653/v1/2020.socialnlp-1.2} {Demoting
  racial bias in hate speech detection}.
\newblock In \emph{Proceedings of the Eighth International Workshop on Natural
  Language Processing for Social Media}, pages 7--14, Online. Association for
  Computational Linguistics.

\bibitem[{Xie et~al.(2017)Xie, Dai, Du, Hovy, and
  Neubig}]{xie_2017_controllable}
Qizhe Xie, Zihang Dai, Yulun Du, Eduard Hovy, and Graham Neubig. 2017.
\newblock Controllable invariance through adversarial feature learning.
\newblock In \emph{Proceedings of the 31st International Conference on Neural
  Information Processing Systems}, pages 585--596.

\bibitem[{Xu et~al.(2021)Xu, Luo, Zhang, Tan, Chang, Huang, and
  Huang}]{xu_2021_child}
Runxin Xu, Fuli Luo, Zhiyuan Zhang, Chuanqi Tan, Baobao Chang, Songfang Huang,
  and Fei Huang. 2021.
\newblock \href {https://doi.org/10.18653/v1/2021.emnlp-main.749} {Raise a
  child in large language model: Towards effective and generalizable
  fine-tuning}.
\newblock In \emph{Proceedings of the 2021 Conference on Empirical Methods in
  Natural Language Processing}, pages 9514--9528, Online and Punta Cana,
  Dominican Republic. Association for Computational Linguistics.

\bibitem[{Zerveas et~al.(2022)Zerveas, Rekabsaz, Cohen, and
  Eickhoff}]{zerveas_2022_mitigating}
George Zerveas, Navid Rekabsaz, Daniel Cohen, and Carsten Eickhoff. 2022.
\newblock \href {https://doi.org/10.1145/3477495.3531891} {Mitigating bias in
  search results through contextual document reranking and neutrality
  regularization}.
\newblock In \emph{Proceedings of the 45th International ACM SIGIR Conference
  on Research and Development in Information Retrieval}, SIGIR '22, page
  2532–2538, New York, NY, USA. Association for Computing Machinery.

\bibitem[{Zhang et~al.(2021)Zhang, van~de Meent, and
  Wallace}]{zhang_2021_disentangling}
Xiongyi Zhang, Jan-Willem van~de Meent, and Byron Wallace. 2021.
\newblock \href {https://doi.org/10.18653/v1/2021.emnlp-main.60} {Disentangling
  representations of text by masking transformers}.
\newblock In \emph{Proceedings of the 2021 Conference on Empirical Methods in
  Natural Language Processing}, pages 778--791, Online and Punta Cana,
  Dominican Republic. Association for Computational Linguistics.

\bibitem[{Zhao et~al.(2019)Zhao, Wang, Yatskar, Cotterell, Ordonez, and
  Chang}]{zhao2019gender}
Jieyu Zhao, Tianlu Wang, Mark Yatskar, Ryan Cotterell, Vicente Ordonez, and
  Kai-Wei Chang. 2019.
\newblock Gender bias in contextualized word embeddings.
\newblock In \emph{Proceedings of the Conference of the North American Chapter
  of the Association for Computational Linguistics: Human Language
  Technologies}, pages 629--634.

\bibitem[{Zhao et~al.(2020)Zhao, Lin, Mi, Jaggi, and
  Sch{\"u}tze}]{zhao_2020_masking}
Mengjie Zhao, Tao Lin, Fei Mi, Martin Jaggi, and Hinrich Sch{\"u}tze. 2020.
\newblock \href {https://doi.org/10.18653/v1/2020.emnlp-main.174} {Masking as
  an efficient alternative to finetuning for pretrained language models}.
\newblock In \emph{Empirical Methods in Natural Language Processing}, pages
  2226--2241. Association for Computational Linguistics.

\bibitem[{Zhou et~al.(2019)Zhou, Lan, Liu, and
  Yosinski}]{zhou2019deconstructing}
Hattie Zhou, Janice Lan, Rosanne Liu, and Jason Yosinski. 2019.
\newblock Deconstructing lottery tickets: Zeros, signs, and the supermask.
\newblock \emph{Advances in Neural Information Processing Systems},
  32:3597--3607.

\end{thebibliography}
\bibliographystyle{acl_natbib}

\clearpage

\appendix

\section{Experiment Settings -- Additional Details}
\label{sec:appendix:setting}

In FDCL18 dataset, we use the TwitterAAE model~\cite{blodgett_2016_demographic} to assign racial dialect classes. The TwiiterAAE model predicts four racial classes, \emph{African American}, \emph{White American}, \emph{Hispanic}, and \emph{Others}. We labeled a tweet as \emph{African American} or \emph{White American} if the prediction score was greater than $0.5$. For PAN16 dataset, following \cite{sap_2019_risk} we balanced the task labels and sampled 200K data. The age groups of this dataset are 18-24, 25-34, 35-49, 50-64, and 65+.

We randomly split the dataset into train, validation, and test set with the proportions 63:12:15 for BIOS, 63:12:15 for FDCL18, and 80:5:15 on PAN16. We use the validation set for hyperparameter tuning, and the best result on the validation set is evaluated on test set for the final results. The validation and test sets in all datasets follow the same distribution as the whole dataset. To address the unbalancedness of the dataset and the potential problems in adversarial learning, we apply upsampling only on the \emph{training sets} of BIOS and FDCL18 datasets, to balance the protected attribute labels within each task label. For instance, genders are balanced in the dentist class by repeating the data items of the minority subgroup. 

\begin{table}[t]
\small
    \begin{center}
    \begin{tabular}{l r} 
    \toprule
    \multicolumn{2}{c}{\textbf{training}}\\
    \midrule
    batch\_size & 64 \\
    structured\_diff\_pruning & True \\
    alpha\_init & 5 \\
    concrete\_samples & 1 \\
    concrete\_lower & -1.5 \\
    concrete\_upper & 1.5 \\
    num\_epochs & 20 \\
    num\_epochs\_finetune & 15 \\
    num\_epochs\_fixmask & 15 \\
    learning\_rate & 2e-05 \\
    learning\_rate\_task\_head & 0.0001 \\
    learning\_rate\_adv\_head & 0.0001 \\
    learning\_rate\_alpha & 0.1 \\
    task\_dropout & 0.3 \\
    task\_n\_hidden & 0 \\
    adv\_dropout & 0.3 \\
    adv\_n\_hidden & 1 \\
    adv\_count & 5 \\
    adv\_lambda & 1.0 \\
    sparsity\_pen & 1.25e-07 \\
    max\_grad\_norm & 1.0 \\
    \midrule
    \midrule
    \multicolumn{2}{c}{\textbf{adv attack}}\\
    \midrule
    batch\_size & 64 \\
    num\_epochs & 40 \\
    learning\_rate & 0.0001 \\
    adv\_n\_hidden & 1 \\
    adv\_count & 5 \\
    adv\_dropout & 0.3 \\
    \bottomrule
    \end{tabular}
    \end{center}
  \caption[Hyperparameters used for training]{Hyperparameters used for training}
  \label{tab:hyper}
\end{table}

\begin{table}[t!]
\small
\setlength\extrarowheight{4pt}
    \begin{center}
    \begin{tabular}{l r} 
    \toprule
    Parameter Name & Size \\
    \midrule
    word\_embeddings & \num[group-separator={,}]{117204480} \\ 
    intermediate.dense & \num[group-separator={,}]{11811840} \\
    output.dense & \num[group-separator={,}]{11800320} \\
    attention.self.query & \num[group-separator={,}]{2952960} \\
    attention.self.key & \num[group-separator={,}]{2952960} \\
    attention.self.value & \num[group-separator={,}]{2952960} \\
    attention.output.dense & \num[group-separator={,}]{2952960} \\
    position\_embeddings & \num[group-separator={,}]{1966080} \\
    output.adapter & \num[group-separator={,}]{590976}\\
    others & \num[group-separator={,}]{7680} \\
    \bottomrule
    \end{tabular}
    \end{center}
  \caption{BERT-Base number of parameters}
  \label{tab:par_sizes}
\end{table}

Adversarial heads consist of five classifiers with different initialization. The loss for the five classifiers is averaged and accuracy is measured via majority vote. All baseline models are trained for 20 epochs. All \modeldiff and \modelmodular model variants are trained for 30 epochs as they need to account for the two phases of \emph{diff} pruning, where a model requires more training to recover its performance after the magnitude pruning step. We fix the learning rate of BERT weights to $2\mathrm{e}{-5}$ and the learning rate for the classifier heads to $1\mathrm{e}{-4}$. We set the batch size to 64 for all experiments. We keep other \emph{diff}-specific hyperparameters the same as suggested by \citet{guo_2021_diffpruning}. Adapter baselines follow \citet{pfeiffer2021adapterfusion} with reduction factor of two. Rest of the hyperparameters are same as the subnetwork-based models. Table \ref{tab:hyper} reports the hyperparameters of our experiments.

\begin{table*}[t]
\centering
    \begin{tabular}{l cccc | cccc}
        \toprule
        \multirow{3}{*}{Model} & \multicolumn{4}{c|}{\textbf{BIOS}} & \multicolumn{4}{c}{\textbf{FCDL18}} \\
         & \multicolumn{2}{c}{Adversarial} & \multicolumn{2}{c|}{MI Reduction} & \multicolumn{2}{c}{Adversarial} & \multicolumn{2}{c}{MI Reduction} \\
        & Task$\uparrow$ & Probe$\downarrow$ & Task$\uparrow$ & Probe$\downarrow$ & Task$\uparrow$ & Probe$\downarrow$ & Task$\uparrow$ & Probe$\downarrow$ \\
        \midrule
        \modelfine & {$82.9_{0.1}$} & {$65.5_{0.4}$} & {$82.9_{0.1}$} & {$65.5_{0.4}$} & {$82.1_{0.2}$} & {$90.3_{0.9}$} & {$82.1_{0.2}$} & {$90.3_{0.9}$} \\
        \modeladapter & {$81.6_{0.2}$} & {$65.7_{0.2}$} & {$81.6_{0.2}$} & {$65.7_{0.2}$} & {$81.9_{0.0}$} & {$79.2_{0.5}$} & {$81.9_{0.0}$} & {$79.2_{0.5}$} \\
        \modelfinedeb & {$81.6_{0.2}$} & {$56.4_{1.7}$} & {$81.6_{0.1}$} & {$59.9_{0.9}$} & {$80.0_{0.5}$} & {$73.7_{2.7}$} & {$79.7_{0.5}$} & {$87.3_{2.6}$} \\
        \modeladapterdeb  &  {$81.5_{0.1}$} & {$63.4_{0.1}$} & {$81.7_{0.1}$} &{$65.4_{0.1}$} & {$81.1_{0.2}$} & {$64.6_{1.3}^{\clubsuit}$} & {$81.8_{0.1}$} &{$78.5_{0.5}^{\clubsuit}$} \\
        \cdashlinelr{1-9}
        \modeldiff & {$83.5_{0.1}$} & {$65.8_{0.3}$} & {$83.5_{0.1}$} & {$65.8_{0.3}$} & {$82.8_{0.1}$} & {$92.6_{0.9}$} & {$82.8_{0.1}$} & {$92.6_{0.9}$} \\
        \modeldiffdeb & {$83.3_{0.1}$} & {$59.1_{0.7}$} & {$82.3_{0.1}$} & {$65.5_{0.9}$} & {$82.3_{0.4}$} & {$65.8_{2.8}$} & {$82.5_{0.3}$} & {$91.9_{0.9}$} \\
        \modelmodularseq & {$83.1_{0.0}$} & {$57.3_{0.7}$} & {$83.1_{0.0}$} & {$63.4_{1.4}$} & {$81.6_{0.4}$} & {$69.7_{1.9}^{\clubsuit}$} & {$82.3._{0.1}$} & {$89.3_{0.4}$} \\
        \modelmodularpar & {$81.5_{0.2}$} & {$\textbf{55.7}_{0.6}^{\clubsuit}$} & {$81.4 _{0.2}$} & {$\textbf{58.8}_{0.8}^{\clubsuit}$} & {$80.2_{0.3}$} & {$73.8_{6.7}$} & {$79.98 _{0.6}$} & {$\textbf{85.4} _{1.6}$} \\
        
        \bottomrule
    \end{tabular}
  \caption{Results of the BIOS and FCDL18 datasets on \textbf{BERT-Mini} with adversarial bias removal and mutual information (MI) reduction. Task performance is measured with accuracy, and bias mitigation with balanced accuracy of the probes. The protected attribute is gender for BIOS, and race-dialect for FCDL18. The results with the best bias mitigation performance (lowest values) among the models that use \emph{diff} subnetworks (lower part of the table) are shown in \textbf{bold}, and among all models with the $\clubsuit$ symbol.}
  \label{tab:results_other_mini}
\end{table*}

\begin{table*}[hbt!]
\centering
    \begin{tabular}{l ccc ccc}
        \toprule
        \multirow{2}{*}{Model} &
        \multicolumn{3}{c}{Adversarial Bias Mitigation} & \multicolumn{3}{c}{MI Reduction} \\
        & Task$\uparrow$ & $\text{Probe}_{\verb|G|}\downarrow$ &  $\text{Probe}_{\verb|A|}$$\downarrow$ & Task$\uparrow$ & $\text{Probe}_{\verb|G|}$$\downarrow$ &  $\text{Probe}_{\verb|A|}$$\downarrow$ \\
        \midrule
        \modelfine & {$91.5_{0.2}$} & {$64.8_{0.6}$} & {$46.8_{0.3}$} & {$91.5_{0.2}$} & {$64.8_{0.3}$} & {$88.4_{1.1}$} \\
        \modeladapter & {$78.4_{0.2}$} & {$65.8_{0.2}$} & {$35.3_{0.8}$} & {$78.4_{0.2}$} & {$65.8_{0.2}$} & {$35.3_{0.8}$} \\
        $\modelfinedeb_{\verb|G|}$ & {$91.7_{0.2}$} & {$54.6_{0.6}$} & {$42.2_{0.6}$} &$91.6_{0.2}$ &$61.3_{0.6}$ &$42.4_{0.5}$ \\
        $\modelfinedeb_{\verb|A|}$ & {$91.1_{0.2}$} & {$61.9_{0.6}$} & {$39.1_{0.8}$} &$91.4_{0.6}$ &$61.9_{0.1}$ &$43.2_{0.0}$ \\
        $\modelfinedeb_{\verb|G&A|}$ & {$91.2_{0.1}$} & {$57.0_{1.0}$} & {$38.5_{0.6}$} &$91.9_{0.0}$ &$62.5_{0.1}$ &$42.1_{0.9}$ \\
        $\modeladapterdeb_{\verb|G|}$ & {$78.1_{0.1}$} & {$59.6_{0.4}$} & {$32.0_{1.3}$} & {$78.5_{0.2}$} & {$65.9_{0.2}$} & {$34.8_{0.0}$} \\
        $\modeladapterdeb_{\verb|A|}$ & {$77.3_{0.1}$} & {$60.5_{0.9}$} & {$27.3_{0.9}$} & {$78.3_{0.1}$} & {$65.5_{0.2}$} & {$34.1_{0.3}^{\clubsuit}$} \\
        $\modeladapterdeb_{\verb|G&A|}$ & {$80.9_{0.6}$} & {$55.6_{0.4}$} & {$25.5_{1.2}^{\clubsuit}$} & {$82.0_{0.1}$} & {$64.2_{0.4}$} & {$34.9_{0.2}$} \\
        \cdashlinelr{1-7}
        \modeldiff & {$90.0_{0.1}$} & {$67.2_{0.3}$} & {$49.4_{0.8}$} & {$90.0_{0.1}$} & {$67.2_{0.3}$} & {$49.4_{0.8}$} \\
        $\modeldiffdeb_{\verb|G|}$ & {$90.1_{0.1}$} & {$\mathbf{54.1}_{1.1}^{\clubsuit}$} & {$44.1_{0.7}$} & {$78.5_{0.2}$} &{$65.9_{0.2}$} &{$34.8_{0.0}$} \\
        $\modeldiffdeb_{\verb|A|}$ & {$89.2_{0.3}$} & {$64.8_{0.6}$} & {$39.4_{1.9}$} & {$78.3_{0.1}$} &{$65.5_{0.2}$} &{$\mathbf{34.1}_{0.3}^{\clubsuit}$} \\
        $\modeldiffdeb_{\verb|G&A|}$ & {$89.1_{0.1}$} & {$57.9_{1.7}$} & {$35.8_{2.2}$} & {$86.3_{0.0}$} &{$68.8_{0.1}$} &{$52.3_{0.9}$} \\
        $\modelmodularseq_{\verb|G|}$ & {$91.7_{0.2}$} & {$55.7_{1.2}$} & {$42.7_{0.6}$} &$91.5_{0.0}$ &$62.6_{0.0}$ &$43.7_{1.3}$ \\
        $\modelmodularseq_{\verb|A|}$ & {$91.4_{0.2}$} & {$62.3_{0.5}$} & {$34.8_{0.8}$} &$91.5_{0.0}$ &$62.8_{0.3}$ & $44.0_{0.0}$ \\
        $\modelmodularseq_{\verb|G|}+\emph{ditto}_{\verb|A|}$ & {$91.0_{0.4}$} & {$59.3_{0.6}$} & {$37.3_{1.2}$} & !!! {$91.5_{0.5}$} & {$62.8_{0.4}$} & {$43.9_{0.8}$} \\
        $\modelmodularseq_{\verb|G&A|}$ & {$91.5_{0.2}$} & {$56.7_{1.3}$} & {$35.1_{2.1}$} &$91.5_{0.0}$ &$63.0_{0.3}$ &$43.9_{0.4}$ \\
        $\modelmodularpar_{\verb|G|}$ & {$91.6_{0.2}$} & {$55.9_{0.8}$} & {$41.7_{0.7}$} &{$91.6_{0.1}$} &{$60.9_{0.4}$} &{$40.3_{0.8}$} \\
        $\modelmodularpar_{\verb|A|}$ & {$91.3_{0.2}$} & {$61.3_{0.5}$} & {$37.6_{1.2}$} &{$91.5_{0.2}$} &{$60.8_{0.6}$} &{$41.6_{0.7}$} \\
        $\modelmodularpar_{\verb|G|}+\emph{ditto}_{\verb|A|}$ & {$91.4_{0.4}$} & {$60.7_{1.0}$} & {$39.6_{1.6}$} &  {$91.7_{0.1}$} & {$60.2_{0.5}^{\clubsuit}$} & {$39.7_{0.1}$} \\
        $\modelmodularpar_{\verb|G&A|}$ & {$91.3_{0.3}$} & {$55.5_{1.1}$} & {$\mathbf{33.7}_{1.6}$} & {$91.5_{0.3}$} &{$\mathbf{60.7}_{1.3}$} &{$41.3_{1.1}$} \\
        
        \bottomrule
    \end{tabular}
  \caption{Results of the PAN16 dataset on \textbf{BERT-Mini}. The subscripts \texttt{G} and \texttt{A} refer to the protected attributes gender and age, respectively. The sign \texttt{G\&A} denotes that the bias mitigation loss of the model consists of the debiasing loss terms of both gender and age. The \modelmodular models in the form of ``$+\emph{ditto}$'' refer to the case, where first two debiasing subnetworks with the same core model are trained separately, and then they are added to the base model at inference time. The results with the best bias mitigation performance (lowest values) among the models that use \emph{diff} subnetworks (lower part of the table) are shown in \textbf{bold}, and among all models with the $\clubsuit$ symbol.}
  \label{tab:results_pan16_mini}
\end{table*}

\begin{table}[h!]
\small
\centering
\begin{subtable}[h]{0.99\columnwidth}
\centering
\begin{tabular}{ l ll | ll}
\toprule
& \multicolumn{2}{c|}{\textbf{BIOS (gender)}} & \multicolumn{2}{c}{\textbf{FDCL18 (race)}} \\
Model & Task$\uparrow$ & Probe$\downarrow$ & Task$\uparrow$ & Probe$\downarrow$   \\
\midrule
BERT-Mini & $71.1_{0.2}$  & $59.8_{0.9}$ & $73.6_{1.6}$ & $57.9_{0.4}$  \\
BERT-Base & $66.2_{0.4}$   & $50.5_{0.3}$ & $71.6_{2.2}$  & $50.2_{0.1}$\\
\bottomrule
\end{tabular}
\caption{BIOS and FDCL18}
\end{subtable}
\bigskip

\begin{subtable}[h]{0.99\columnwidth}
\centering
\begin{tabular}{l l ccc }
\toprule
Model & & Task$\uparrow$ & $\text{Probe}_{G}\downarrow$  & $\text{Probe}_{A}\downarrow$ \\
\midrule
\multirow{2}{*}{BERT-Mini} & ${G}$ & $69.3_{0.1}$ &  $60.1_{0.3}$ & $29.2_{0.8}$ \\
& ${A}$ & $66.7_{1.8}$ & $60.6_{0.1}$ & $25.6_{0.2}$ \\
\multirow{2}{*}{BERT-Base} & ${G}$ & $69.4_{0.1}$ & $54.4_{0.1}$ & $25.5_{0.2}$ \\
& ${A}$ & $49.8_{0.8}$ & $54.6_{0.6}$ & $27.5_{0.2}$\\
\bottomrule
\end{tabular}
\caption{PAN16}
\end{subtable}
\caption{Evaluation results of \modelinlp.}
\label{tbl:results:inlp}
\end{table}


\section{Additional Results}
\label{sec:appendix:results}
The results of all models using BERT-Mini are shown in Table~\ref{tab:results_other_mini} for BIOS and FCDL18 datasets, and in Table~\ref{tab:results_pan16_mini} for PAN16. Table~\ref{tbl:results:inlp} reports the evaluation results of the \modelinlp method.

\section{Sparsity rates of subnetworks}
\label{sec:appendix:nonzero}
\begin{figure*}[t]
  \centering
  
  \begin{subfigure}[t]{0.48\textwidth}
  \centering
  \includegraphics[width=1\textwidth]{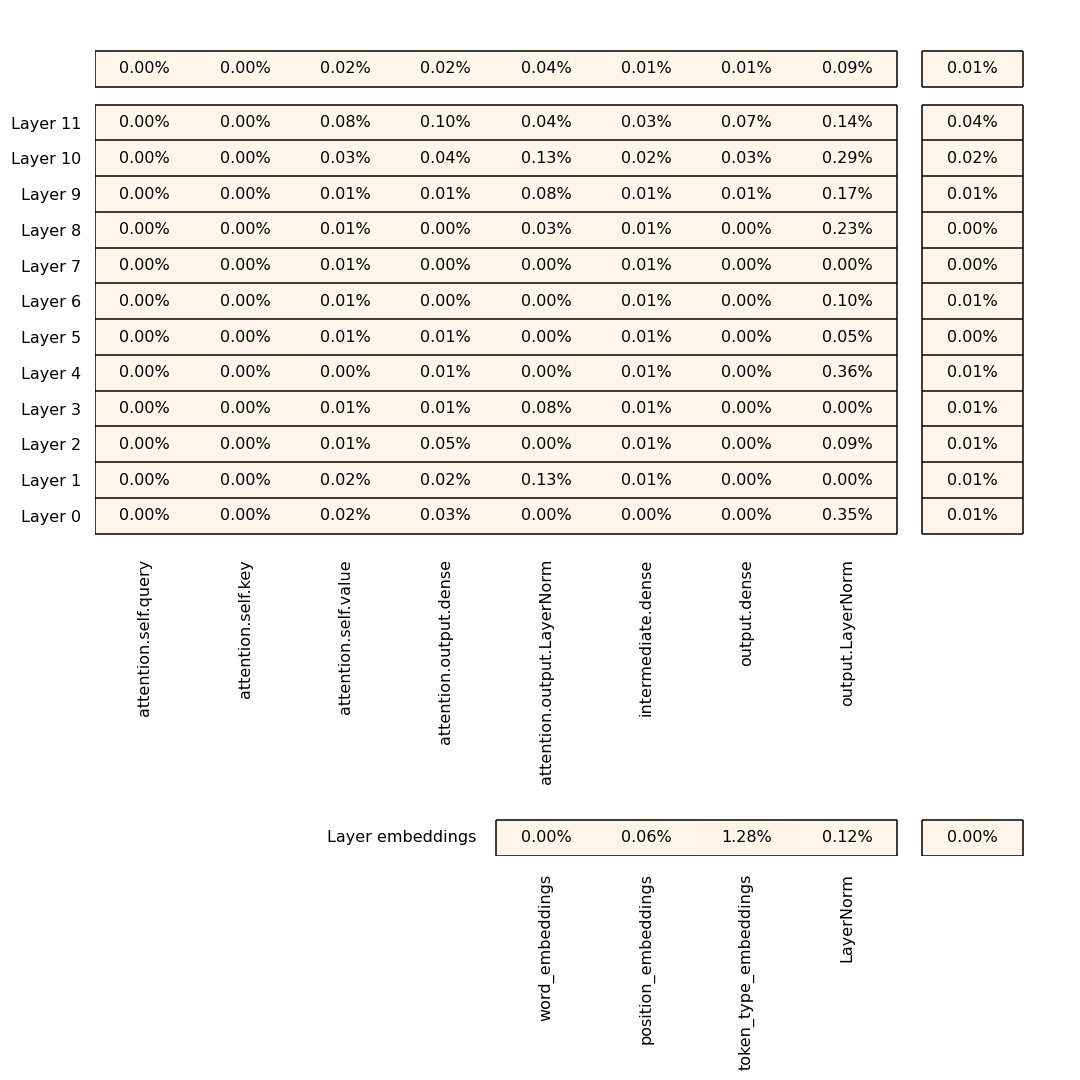}
  \caption{PAN16 - Gender}
  \end{subfigure}
  \hfill
  \begin{subfigure}[t]{0.48\textwidth}
  \centering
  \includegraphics[width=1\textwidth]{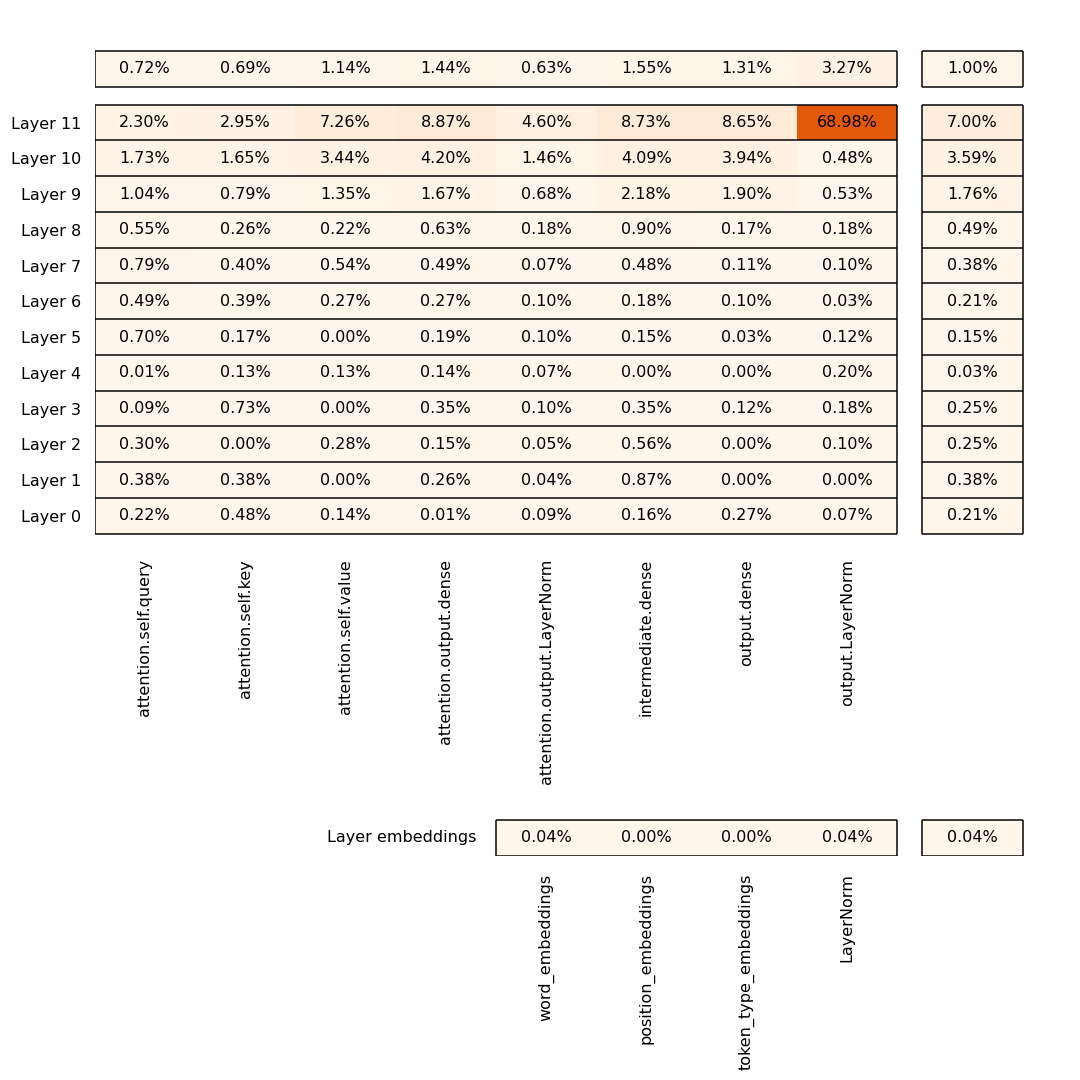}
  \caption{PAN16 - Age}
  \end{subfigure}
  
  \begin{subfigure}[t]{0.48\textwidth}
  \centering
  \includegraphics[width=1\textwidth]
  {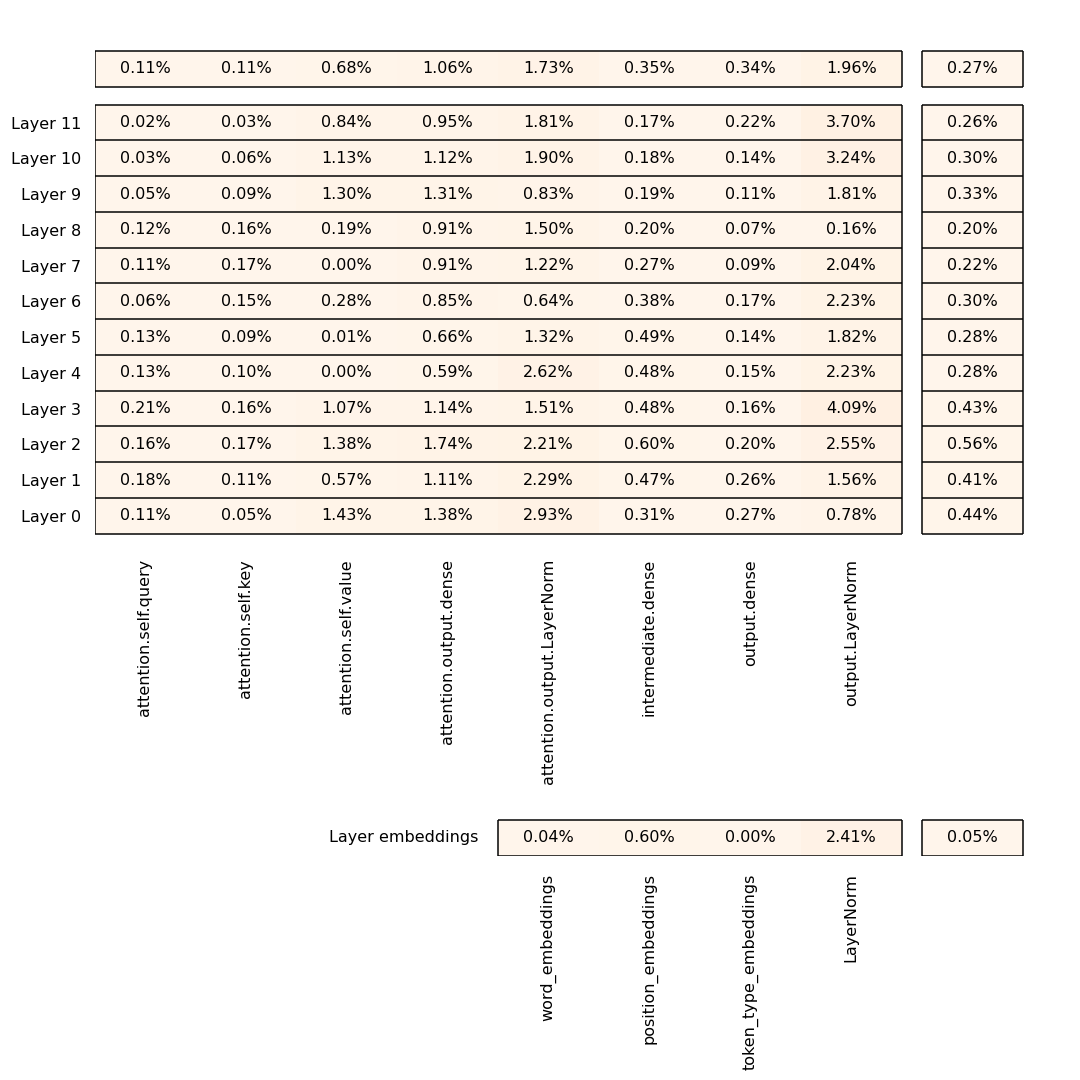}
  \caption{BIOS - Gender}
  \end{subfigure}
  \hfill
  \begin{subfigure}[t]{0.48\textwidth}
  \centering
  \includegraphics[width=1\textwidth]
  {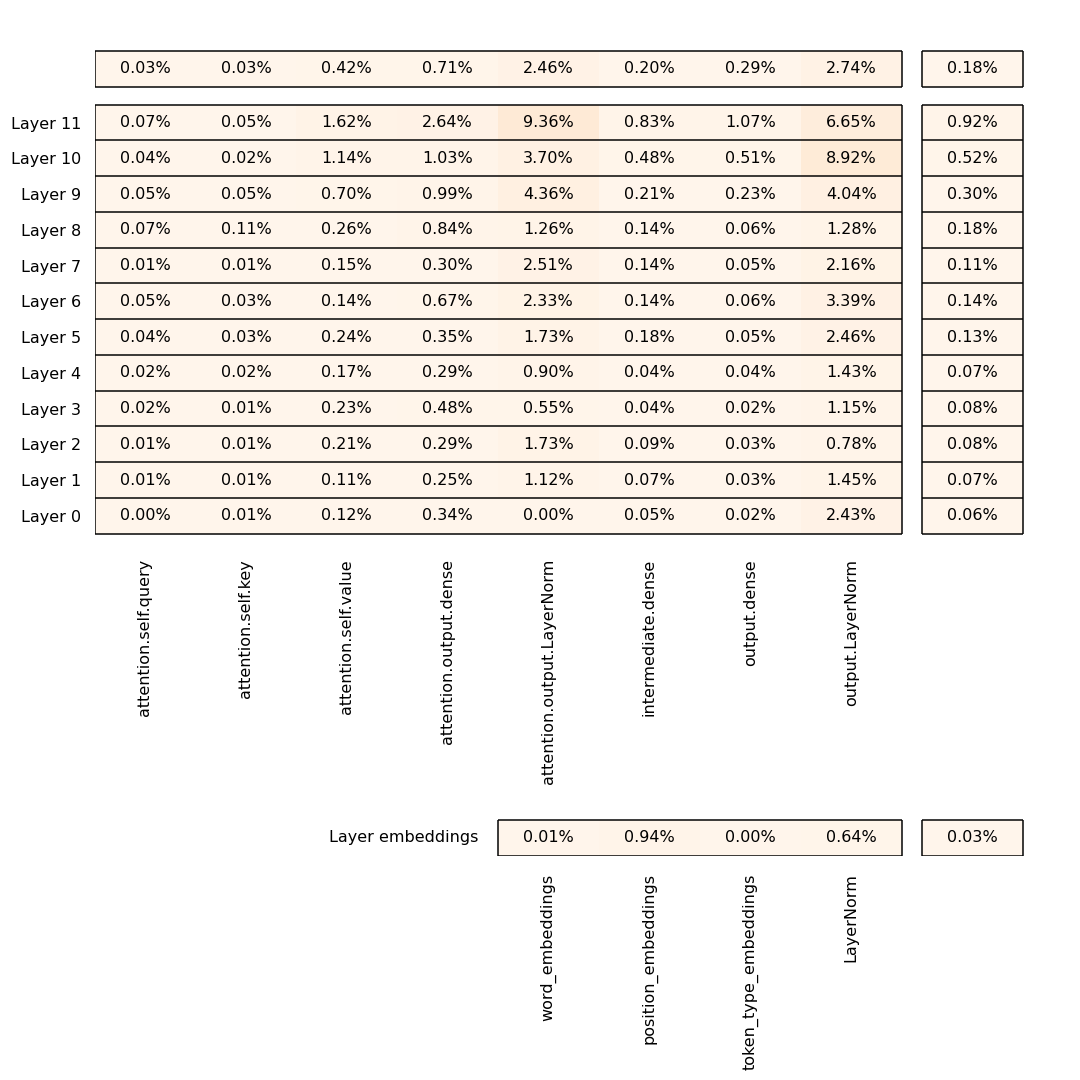}
  \caption{FCDL18 - Dialect}
  \end{subfigure}
  
  \caption{Sparsity rate of the subnetworks in \modelmodularpar on BERT-Base. Each value shows the sparsity rate on the specific block/matrix.}
  \label{fig:density_per_module_modularpar}
\end{figure*}

\begin{figure*}[t]
  \centering
  
  \begin{subfigure}[t]{0.48\textwidth}
  \centering
  \includegraphics[width=1\textwidth]{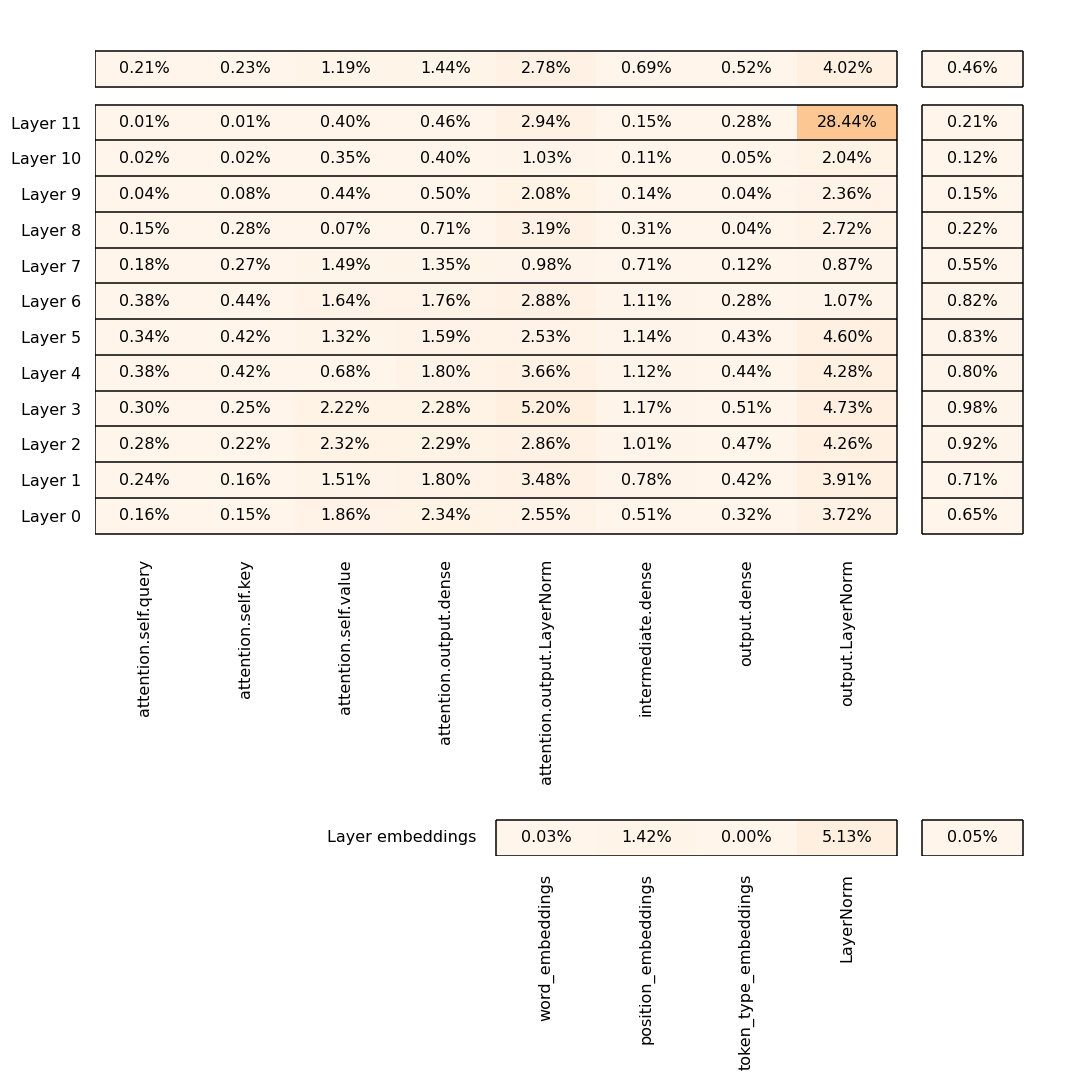}
  \caption{PAN16 - Gender}
  \end{subfigure}
  \hfill
  \begin{subfigure}[t]{0.48\textwidth}
  \centering
  \includegraphics[width=1\textwidth]{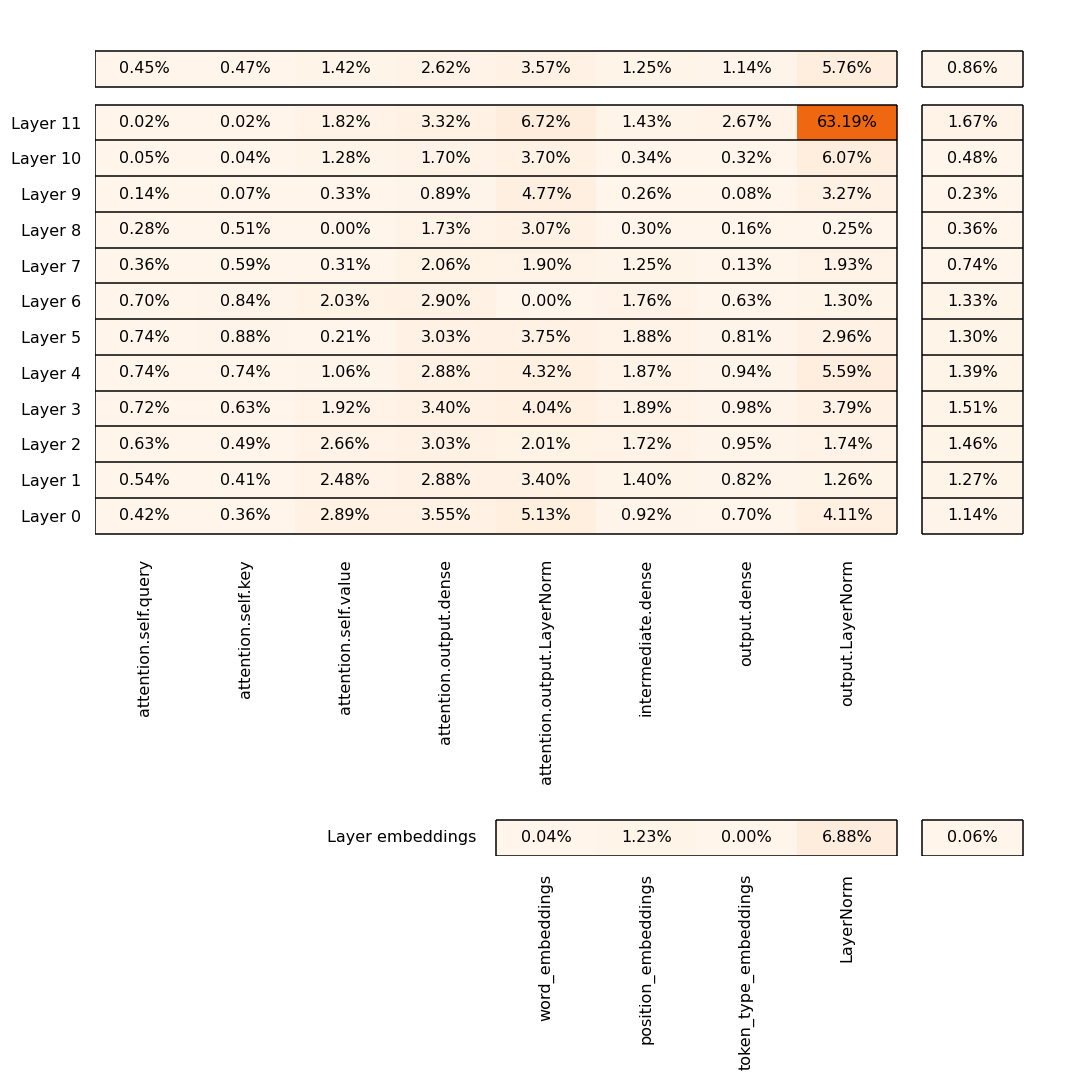}
  \caption{PAN16 - Age}
  \end{subfigure}

  \begin{subfigure}[t]{0.48\textwidth}
  \centering
  \includegraphics[width=1\textwidth]
  {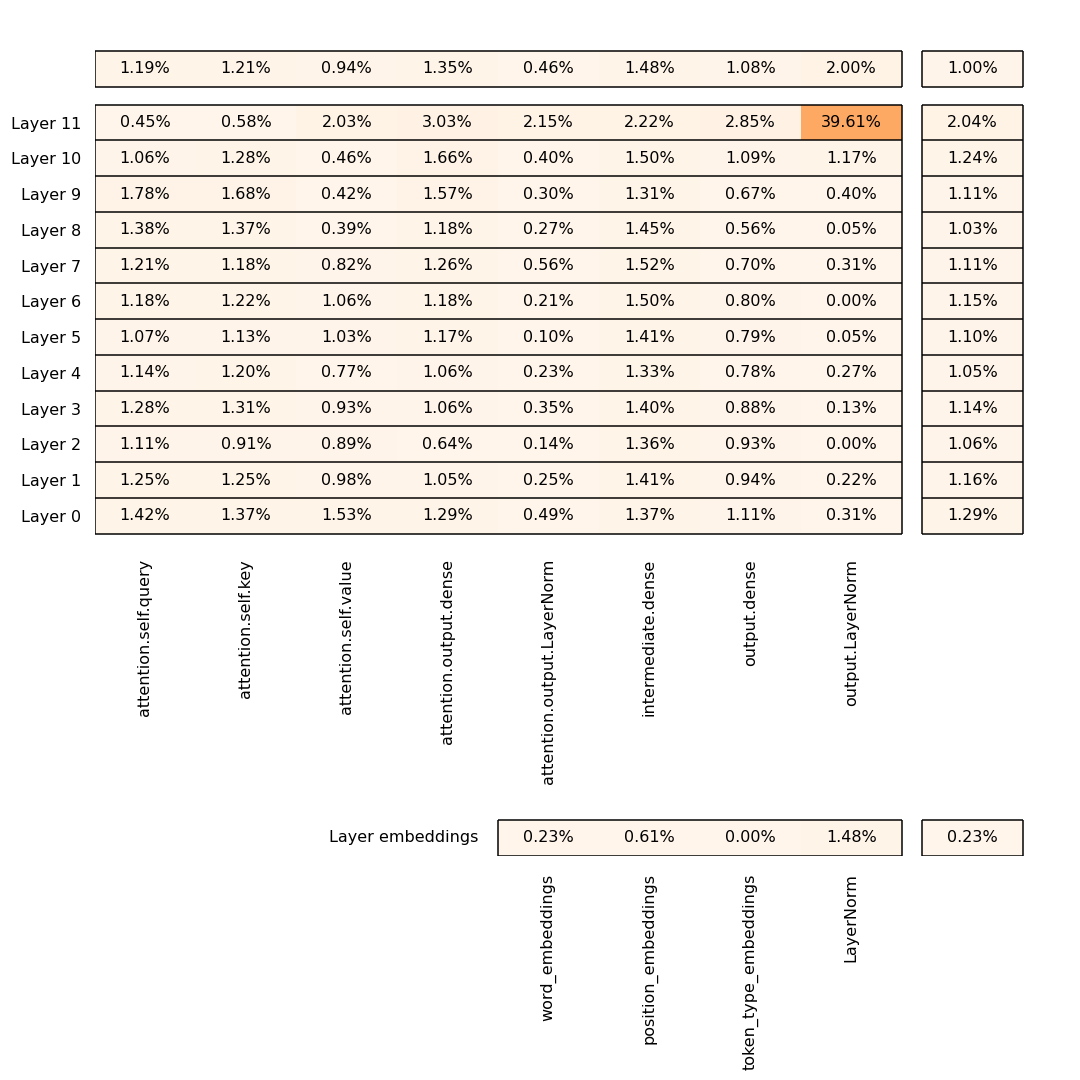}
  \caption{BIOS - Gender}
  \end{subfigure}
  \hfill
  \begin{subfigure}[t]{0.48\textwidth}
  \centering
  \includegraphics[width=1\textwidth]
  {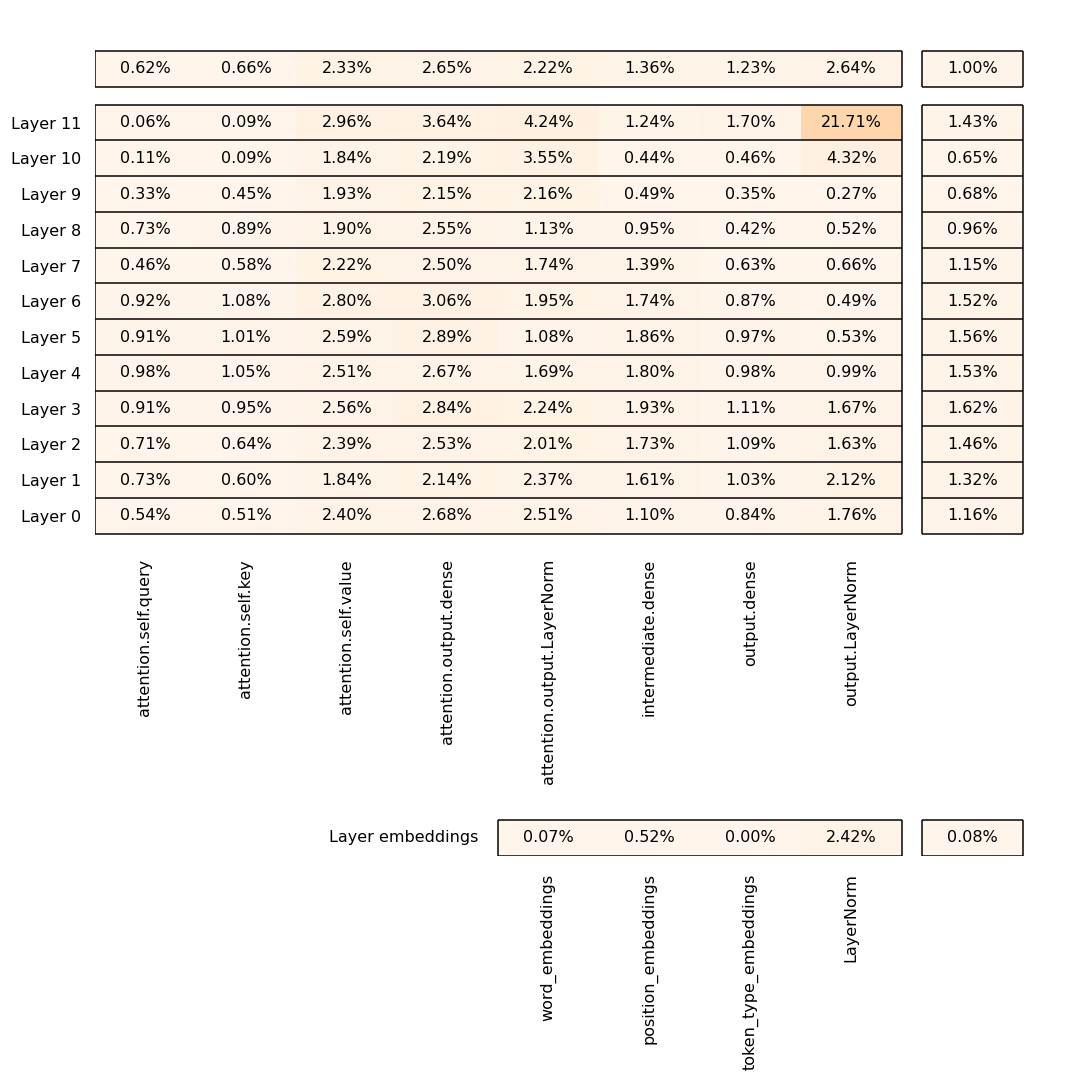}
  \caption{FCDL18 - Dialect}
  \end{subfigure}
  
  \caption{Sparsity rate of the subnetworks in \modelmodularseq on BERT-Base. Each value shows the sparsity rate on the specific block/matrix.}
  \label{fig:density_per_module_modularseq}
\end{figure*}

\begin{figure*}[t]
  \centering
  \begin{subfigure}[t]{0.48\textwidth}
  \centering
  \includegraphics[width=1\textwidth]{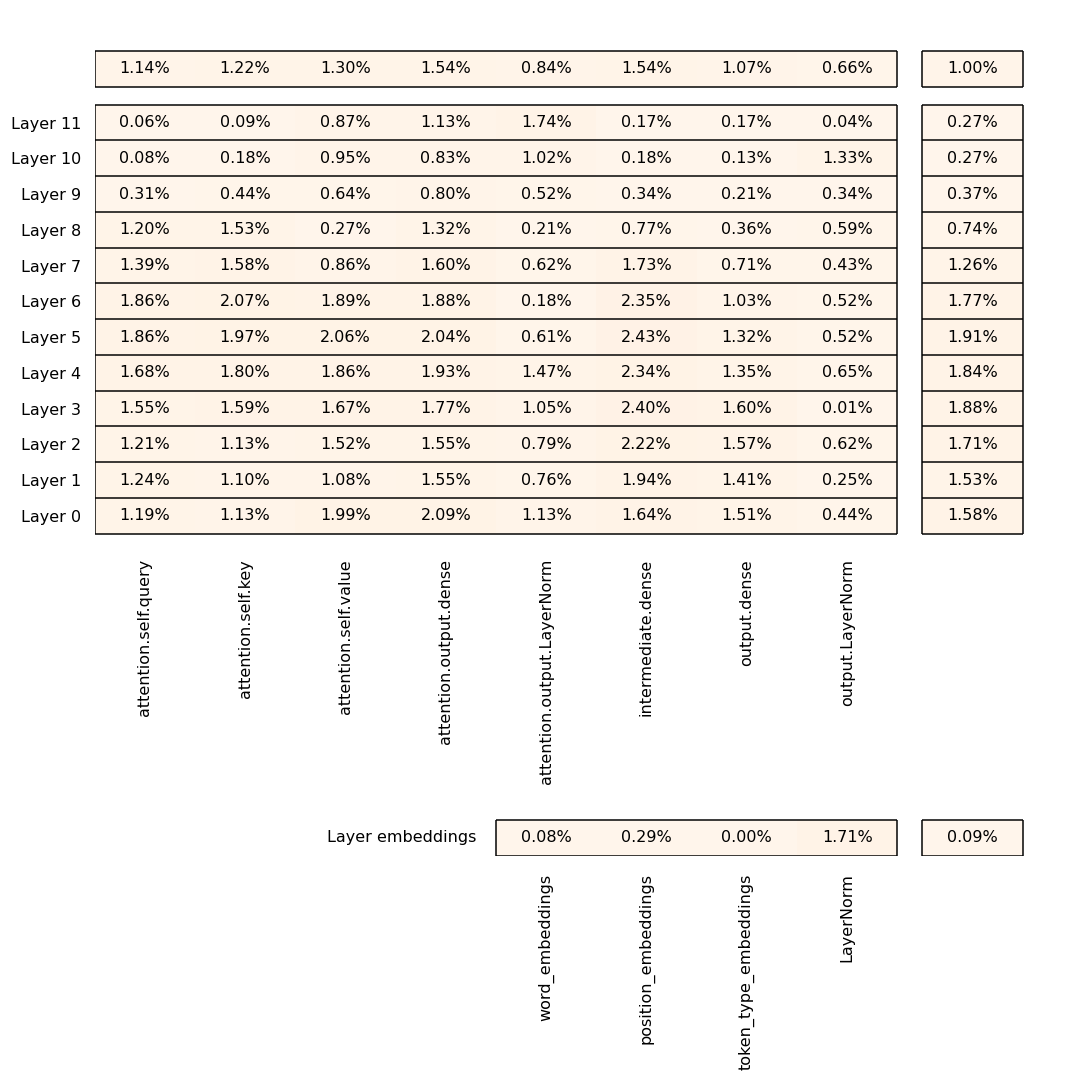}
  \caption{PAN16 - Gender}
  \end{subfigure}
  \hfill
  \begin{subfigure}[t]{0.48\textwidth}
  \centering
  \includegraphics[width=1\textwidth]{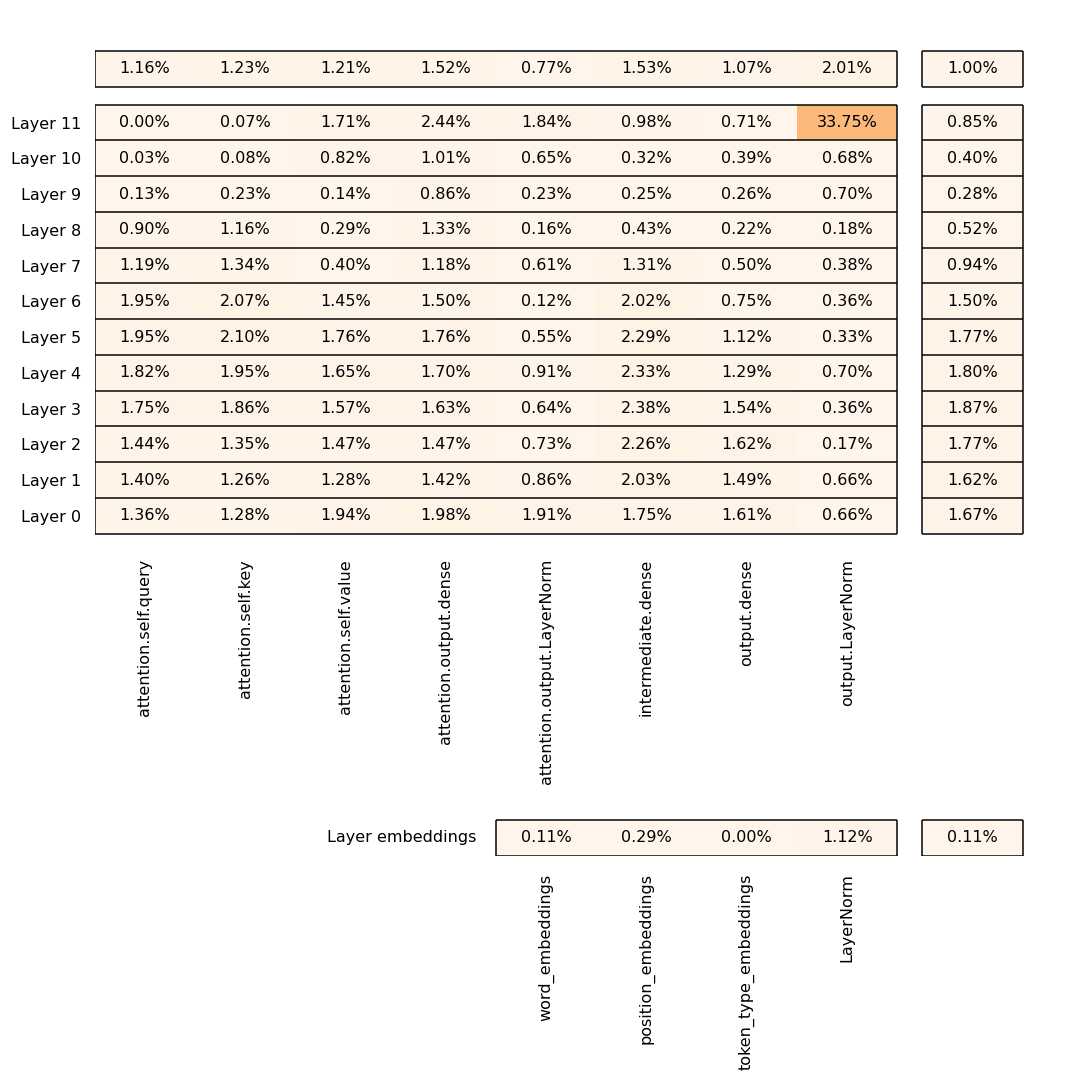}
  \caption{PAN16 - Age}
  \end{subfigure}

  \begin{subfigure}[t]{0.48\textwidth}
  \centering
  \includegraphics[width=1\textwidth]
  {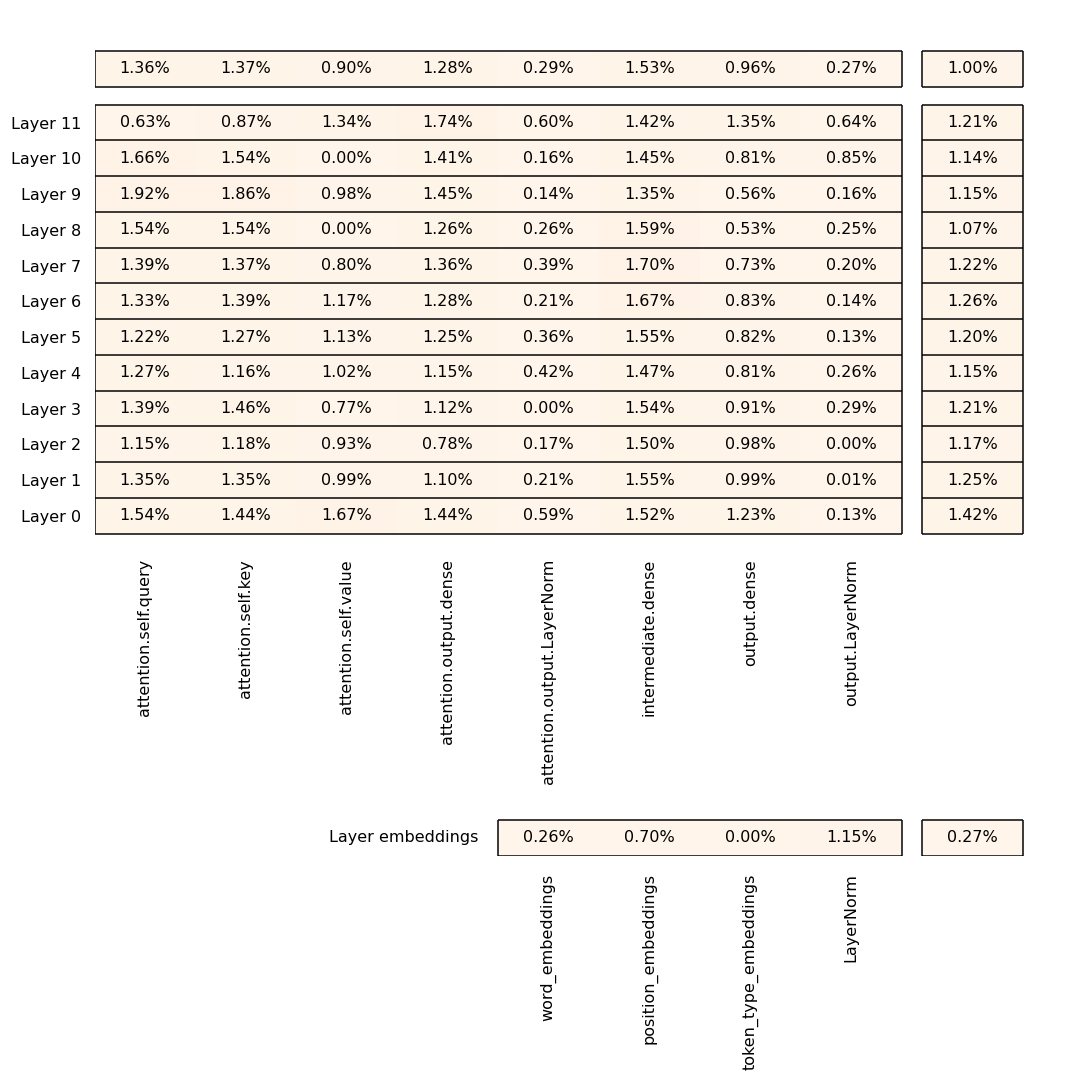}
  \caption{BIOS - Gender}
  \end{subfigure}
  \hfill
  \begin{subfigure}[t]{0.48\textwidth}
  \centering
  \includegraphics[width=1\textwidth]
  {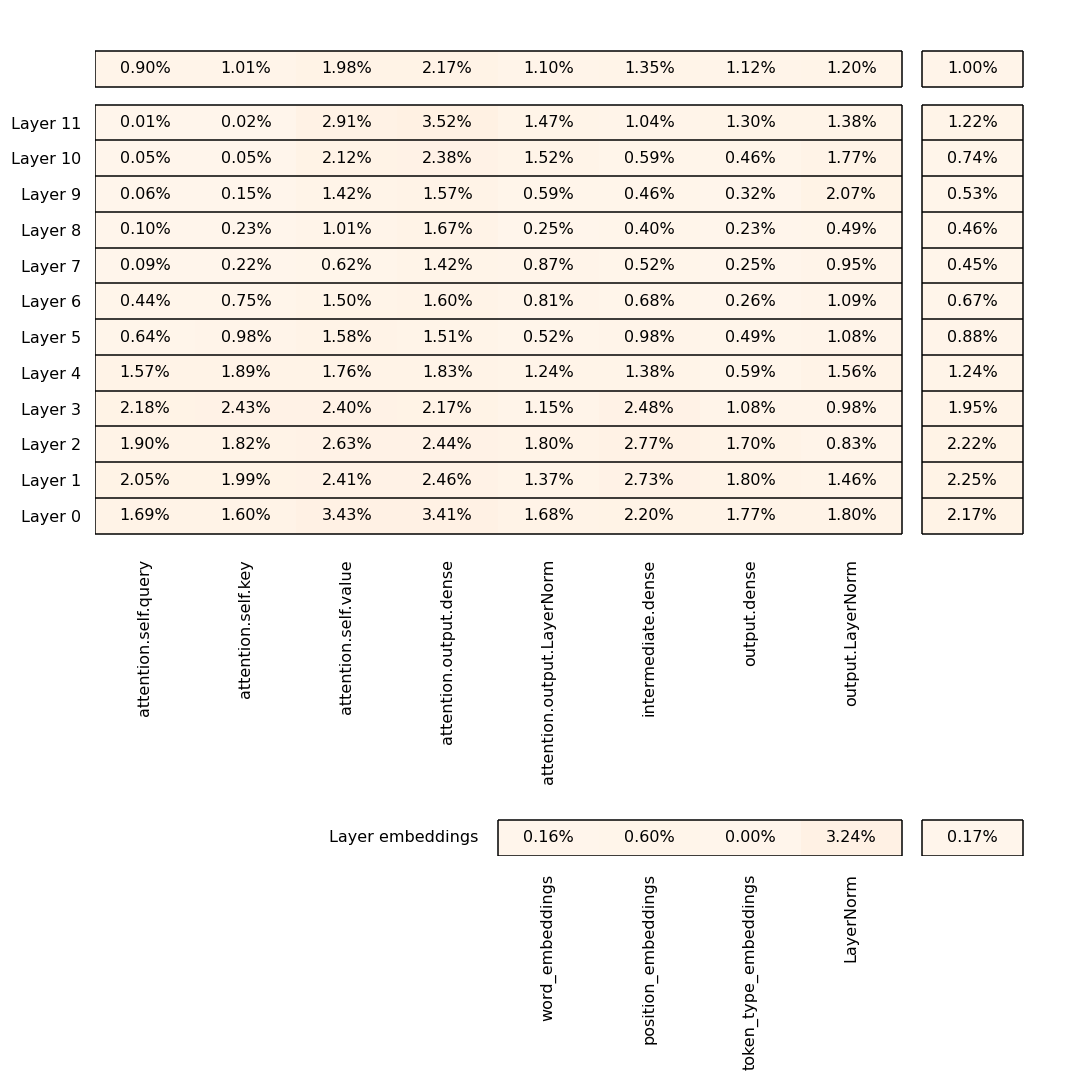}
  \caption{FCDL18 - Dialect}
  \end{subfigure}
  
  \caption{Sparsity rate of the subnetworks in \modeldiffdeb using BERT-Base. Each value shows the sparsity rate on the specific block/matrix.}
  \label{fig:density_per_module_diffdeb}
\end{figure*}

We visualize the percentage of non-masked parameters of the subnetworks in BERT-Base for each parameter matrix in Figures \ref{fig:density_per_module_modularpar}, \ref{fig:density_per_module_modularseq} and \ref{fig:density_per_module_diffdeb} for \modelmodularpar, \modelmodularseq, and \modeldiffdeb, respectively. In addition to the discussion in Section~\ref{sec:results}, we observe in these detailed figures that the LayerNorm module of the last Transformer block generally has a high density. We assume that this is due to the additive nature of these \emph{diff}-based methods, as changing the weight magnitude through adding a subnetwork requires rescaling the final output. 



\section{Consistency in finding subnetworks}
\label{sec:appendix:homogeneity}
\begin{figure*}[t]
  \centering
  \begin{subfigure}[t]{0.66\textwidth}
  \centering
  \includegraphics[width=1\textwidth]{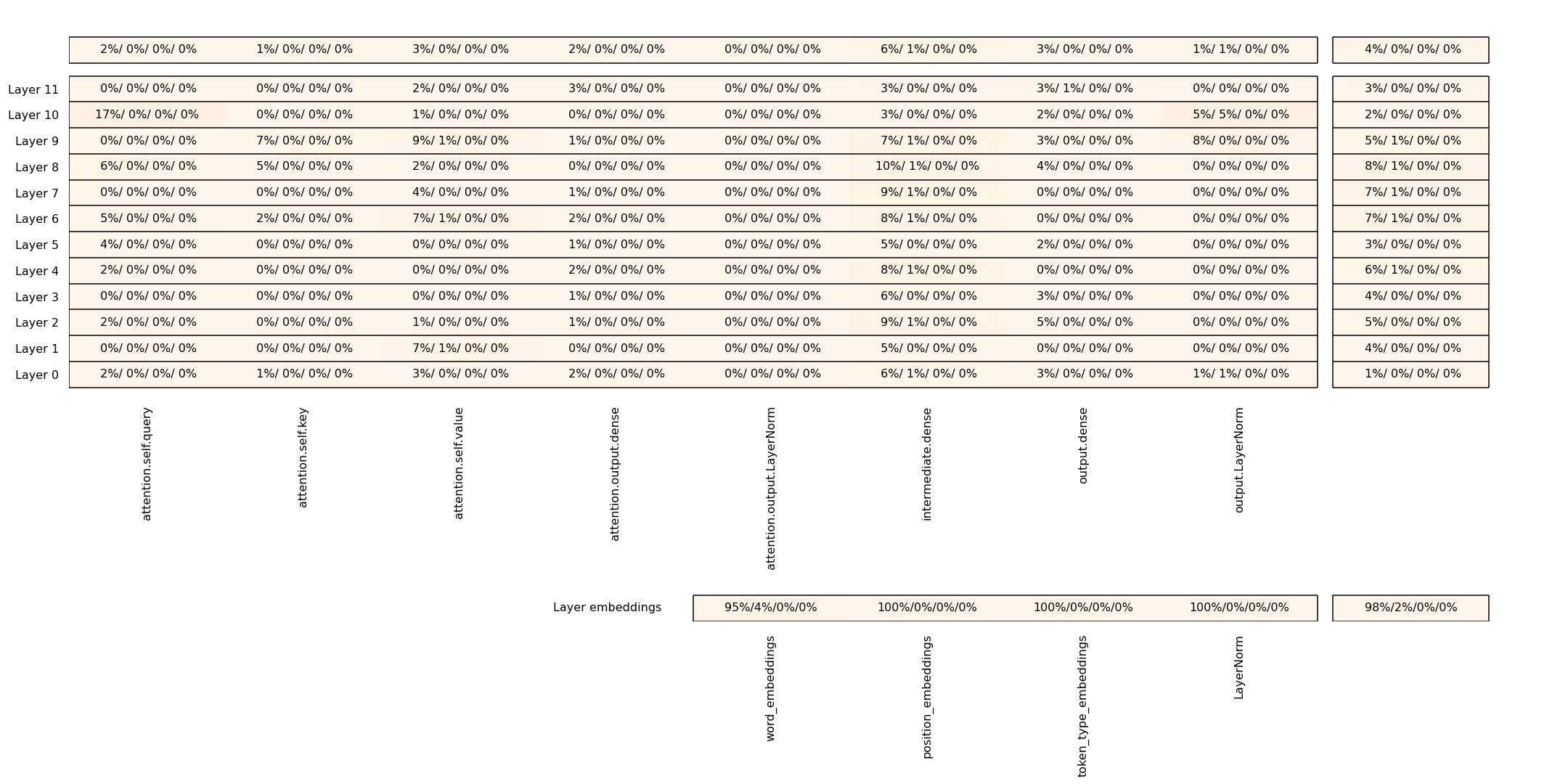}
  \caption{PAN16 - Gender}
  \end{subfigure}
  
  \begin{subfigure}[t]{0.66\textwidth}
  \centering
  \includegraphics[width=1\textwidth]{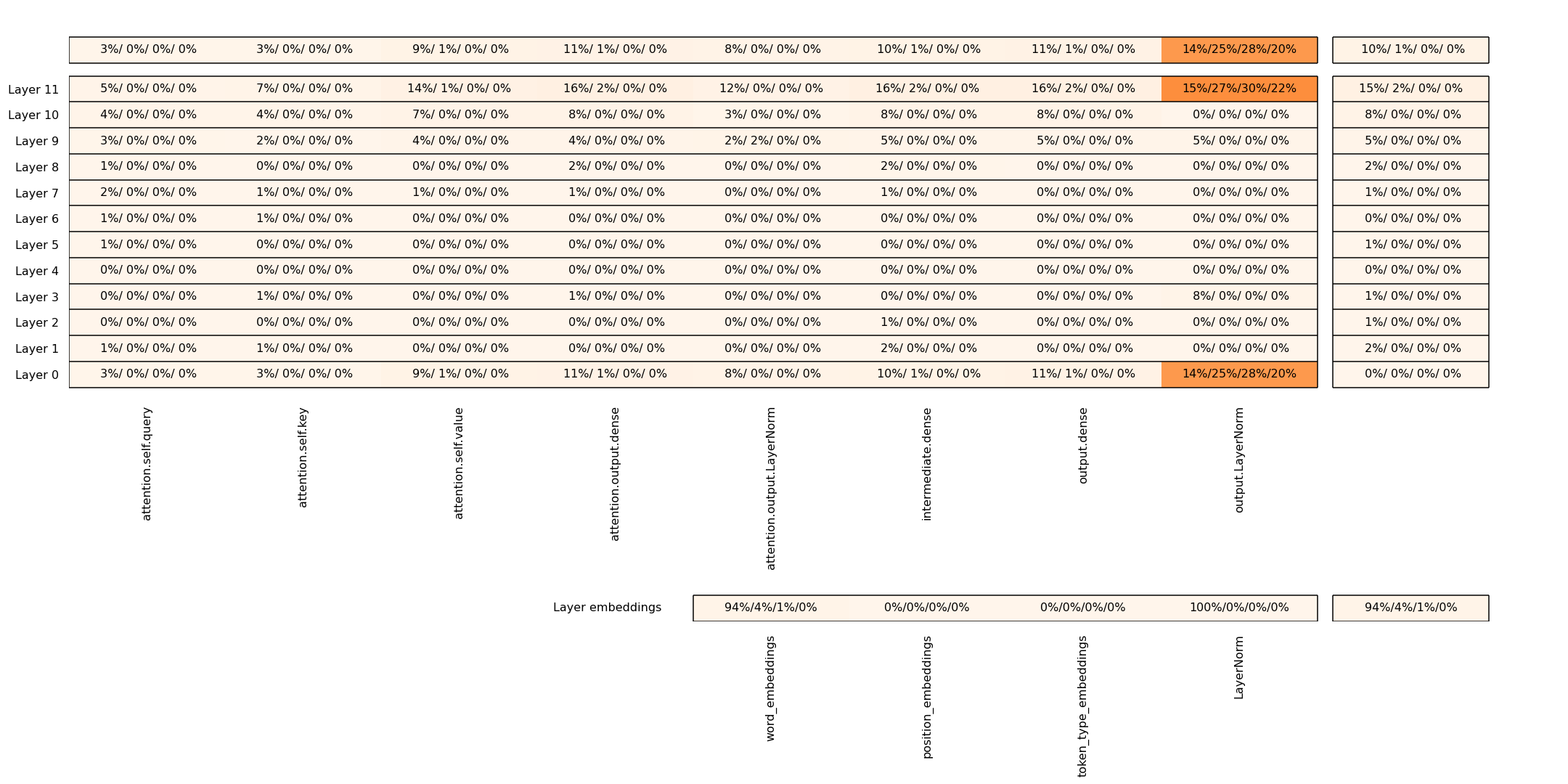}
  \caption{PAN16 - Age}
  \end{subfigure}

  \begin{subfigure}[t]{0.66\textwidth}
  \centering
  \includegraphics[width=1\textwidth]{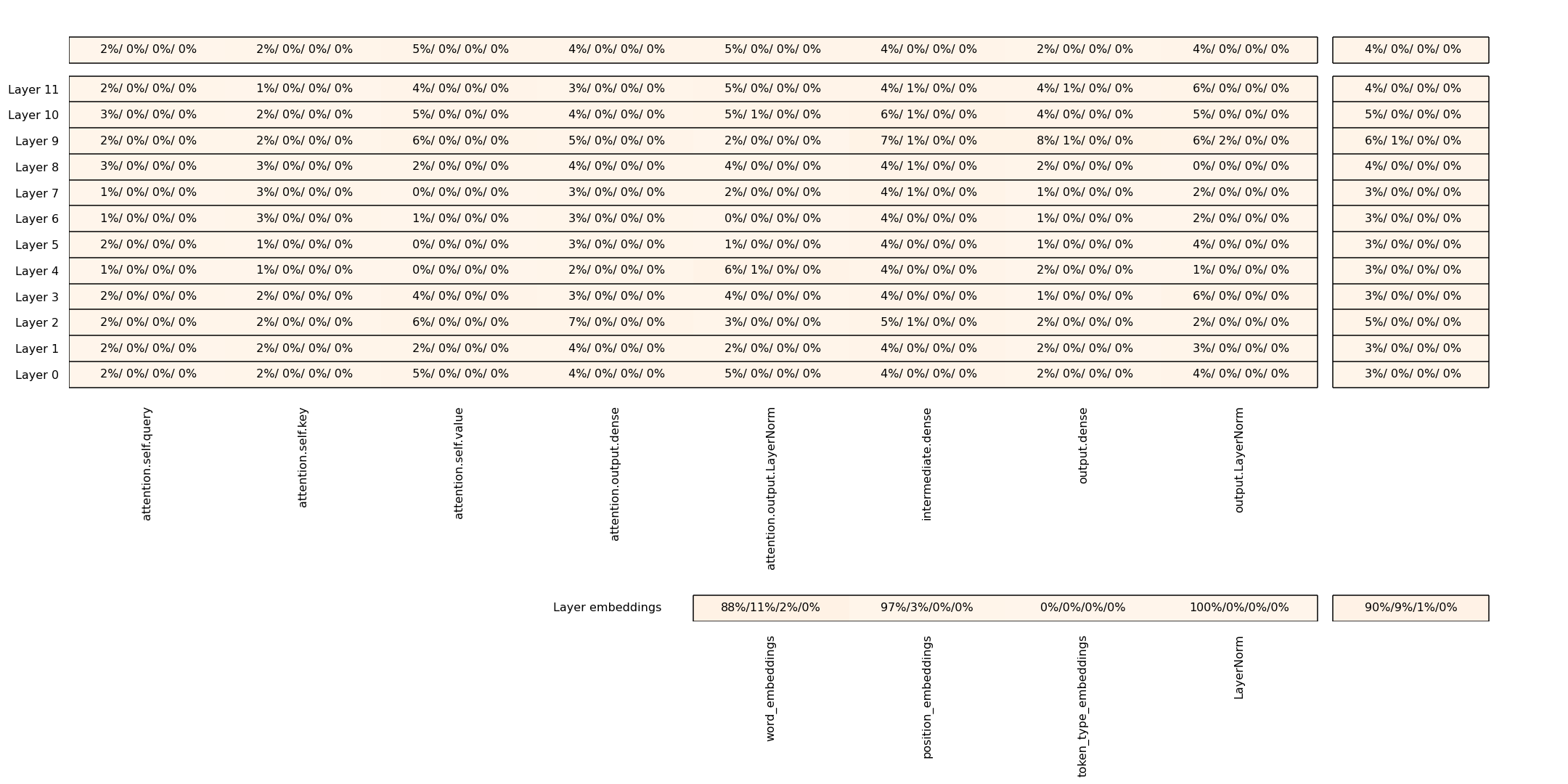}
  \caption{BIOS - Gender}
  \end{subfigure}
  
  \begin{subfigure}[t]{0.66\textwidth}
  \centering
  \includegraphics[width=1\textwidth]{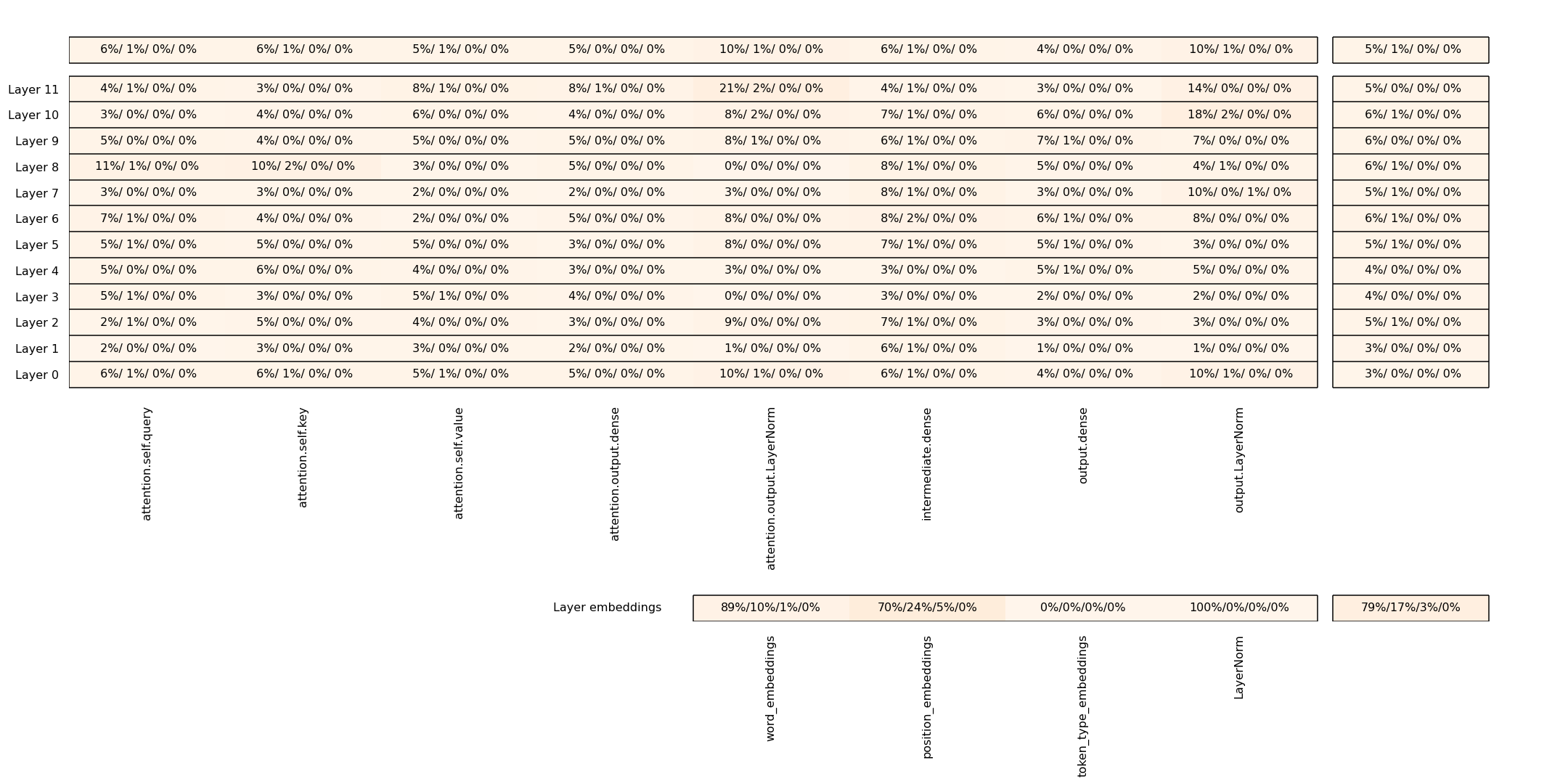}
  \caption{FCDL18 - Dialect}
  \end{subfigure}
  
  \caption{The percentage of common non-masked parameters for \modelmodularpar across 5 runs with different initialization seeds. For each block, the numbers indicate the percentage of the number of common parameters across two, three, four, and five runs of a subnetwork, respectively.}
  \label{fig:overlap_mod}
\end{figure*}

\begin{figure*}[t]
  \centering
  \begin{subfigure}[t]{0.66\textwidth}
  \centering
  \includegraphics[width=1\textwidth]{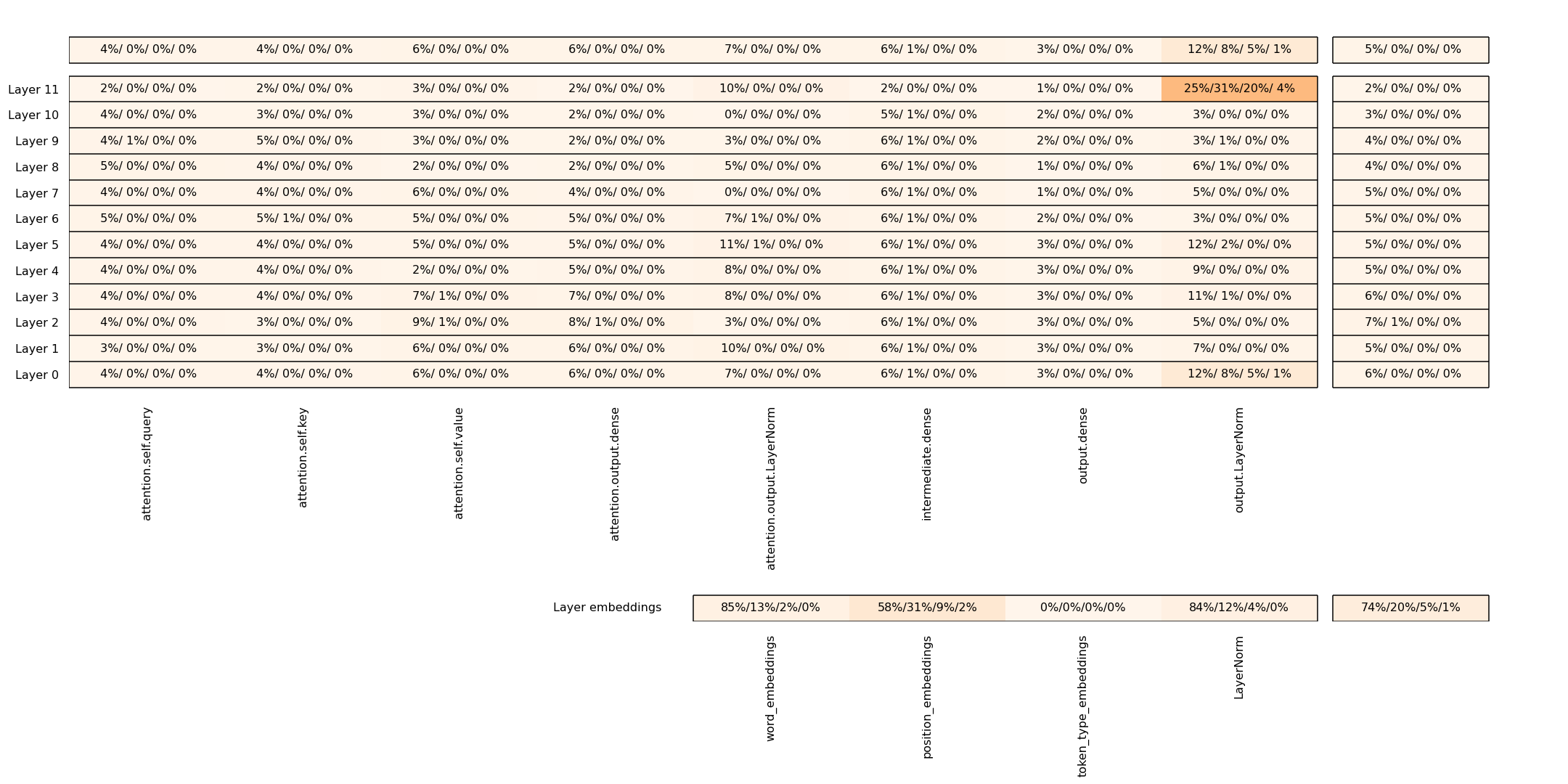}
  \caption{PAN16 - Gender}
  \end{subfigure}
  
  \begin{subfigure}[t]{0.66\textwidth}
  \centering
  \includegraphics[width=1\textwidth]{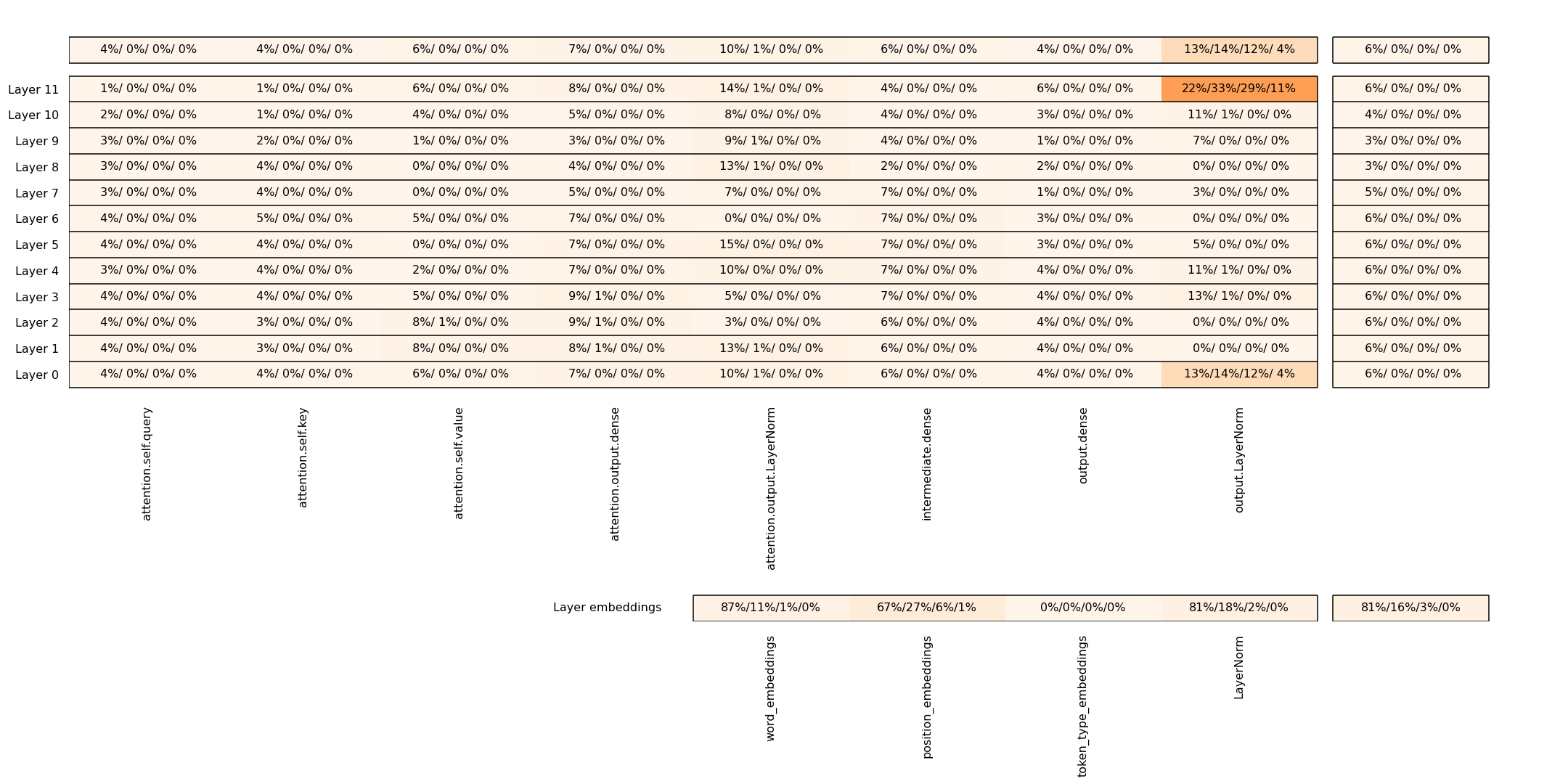}
  \caption{PAN16 - Age}
  \end{subfigure}

  \begin{subfigure}[t]{0.66\textwidth}
  \centering
  \includegraphics[width=1\textwidth]{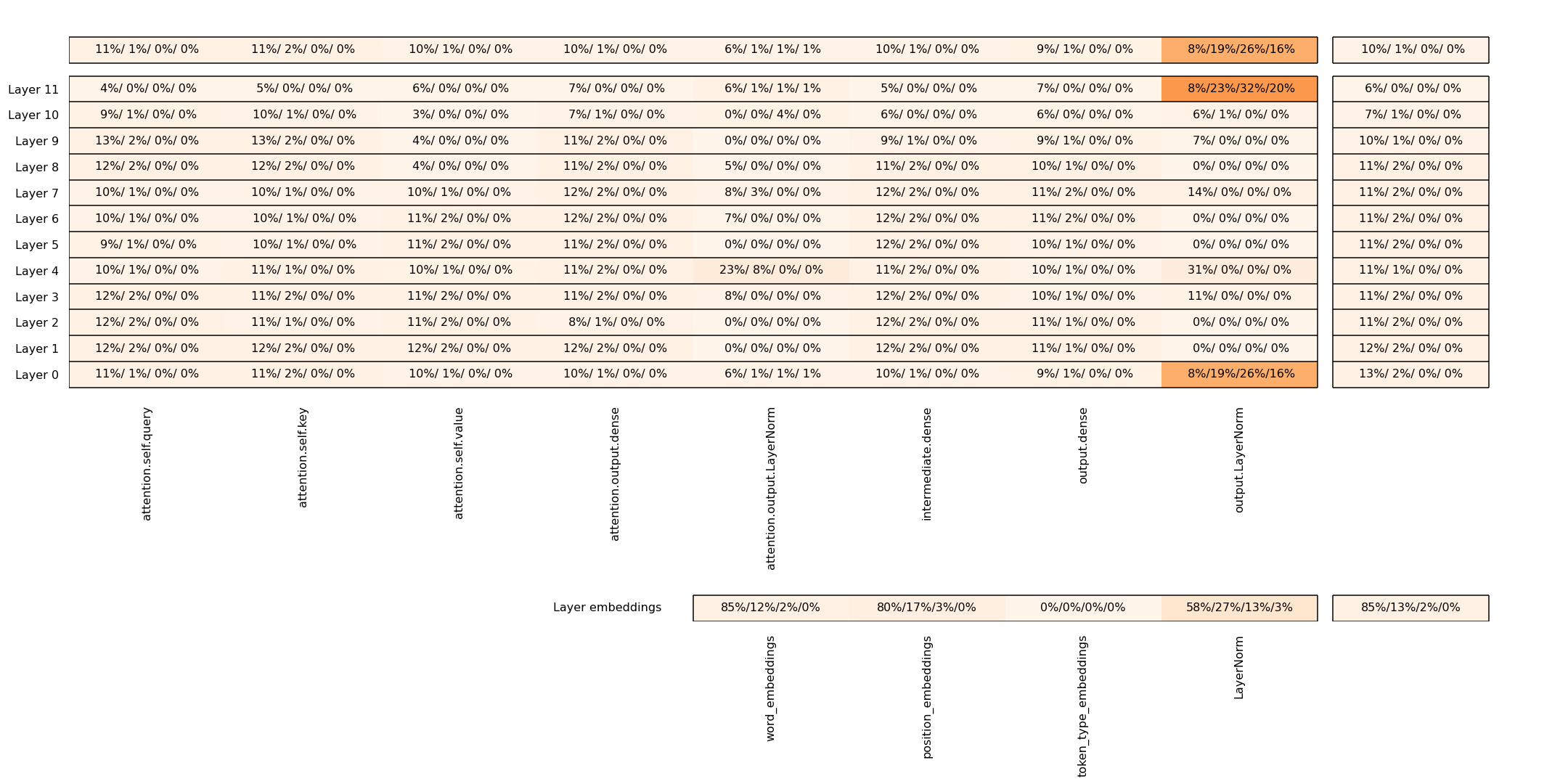}
  \caption{BIOS - Gender}
  \end{subfigure}
  
  \begin{subfigure}[t]{0.66\textwidth}
  \centering
  \includegraphics[width=1\textwidth]{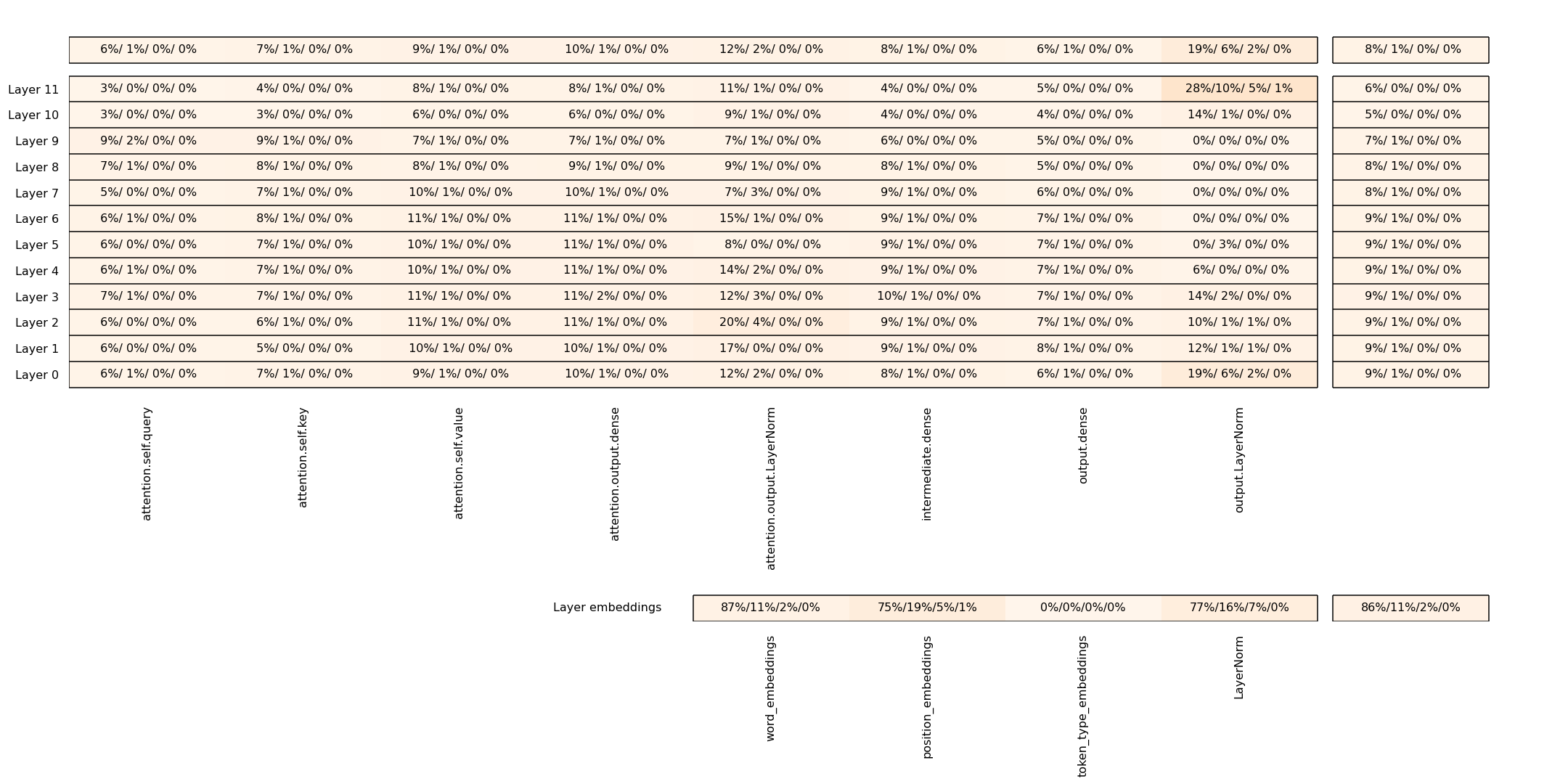}
  \caption{FCDL18 - Dialect}
  \end{subfigure}
  \caption{The percentage of common non-masked parameters for \modelmodularseq across 5 runs with different initialization seeds. For each block, the numbers indicate the percentage of the number of common parameters across two, three, four, and five runs of a subnetwork, respectively.}
  \label{fig:overlap_modseq}
\end{figure*}

\begin{figure*}[t]
  \centering
  
  \begin{subfigure}[t]{0.66\textwidth}
  \centering
  \includegraphics[width=1\textwidth]{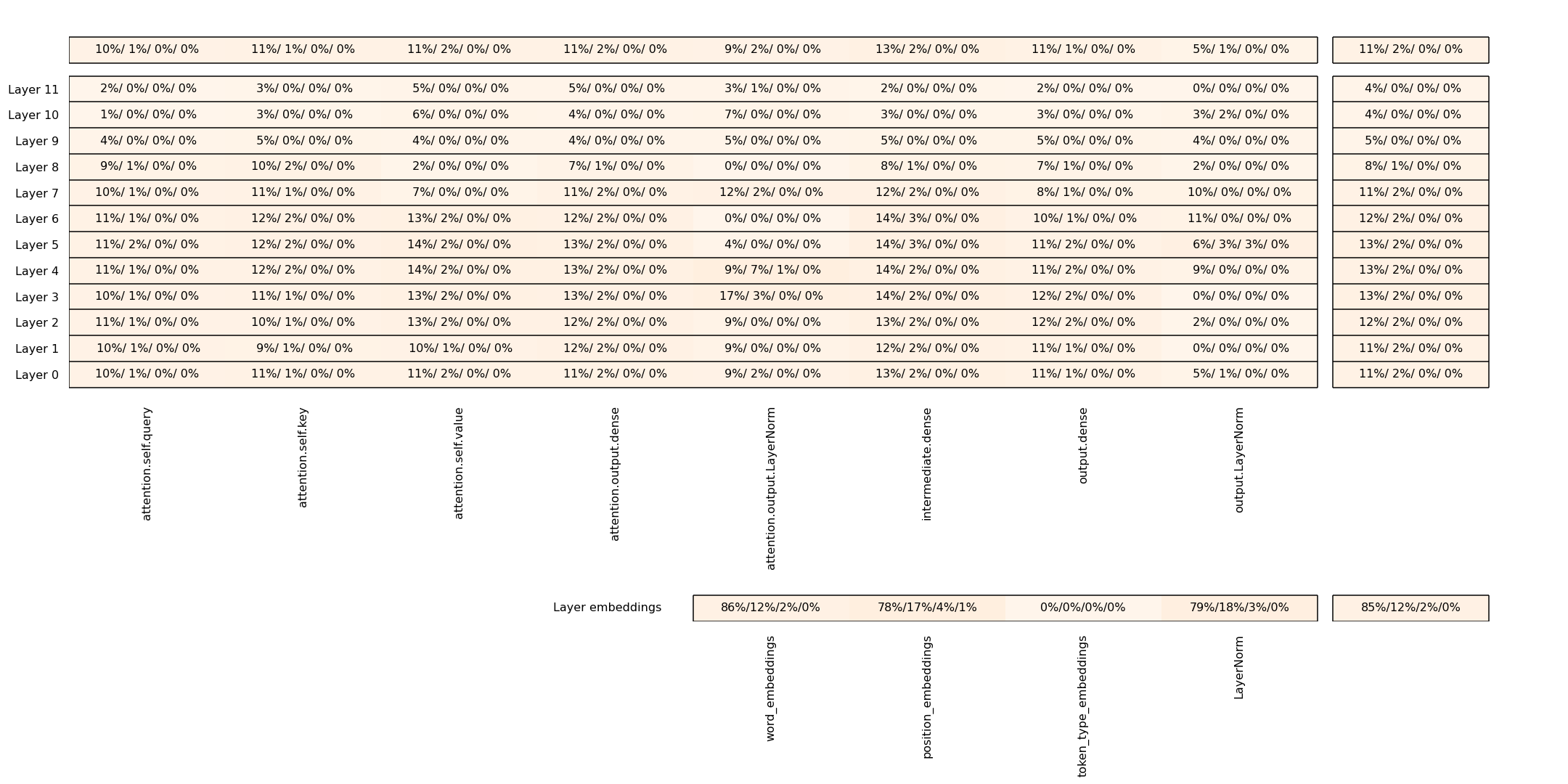}
  \caption{PAN16 - Gender}
  \end{subfigure}
  
  \begin{subfigure}[t]{0.66\textwidth}
  \centering
  \includegraphics[width=1\textwidth]{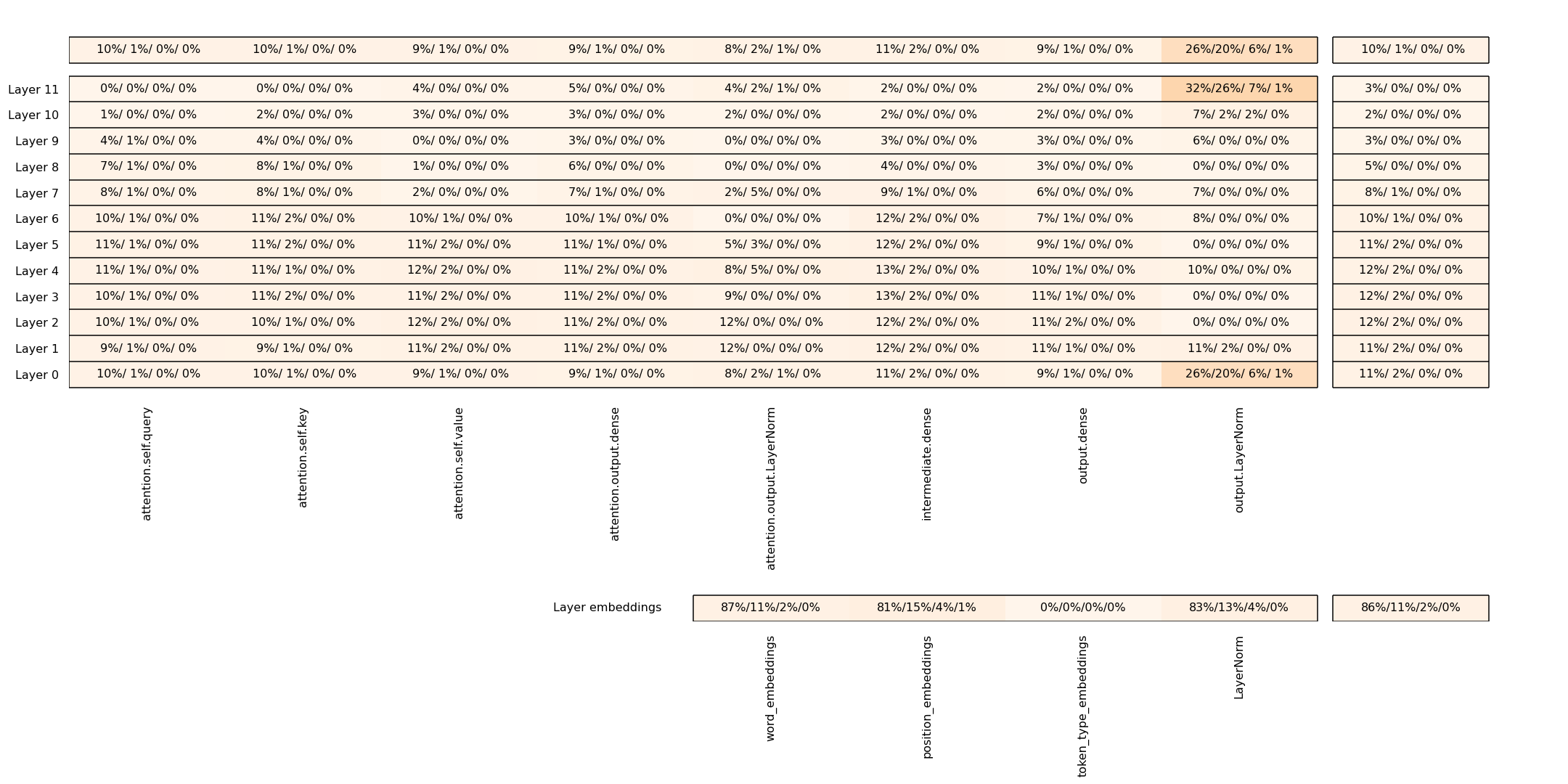}
  \caption{PAN16 - Age}
  \end{subfigure}

  \begin{subfigure}[t]{0.66\textwidth}
  \centering
  \includegraphics[width=1\textwidth]{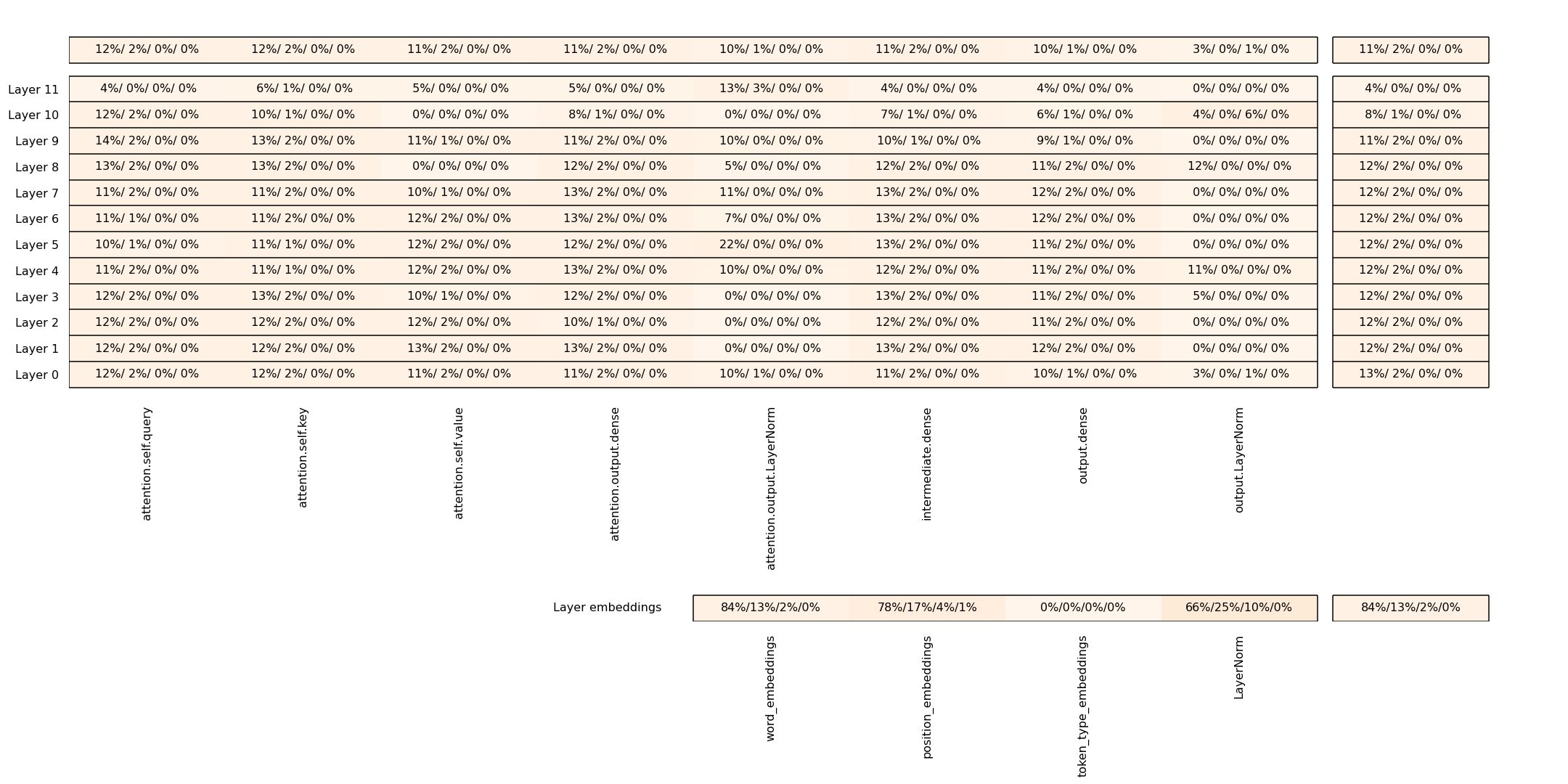}
  \caption{BIOS - Gender}
  \end{subfigure}
  
  \begin{subfigure}[t]{0.66\textwidth}
  \centering
  \includegraphics[width=1\textwidth]{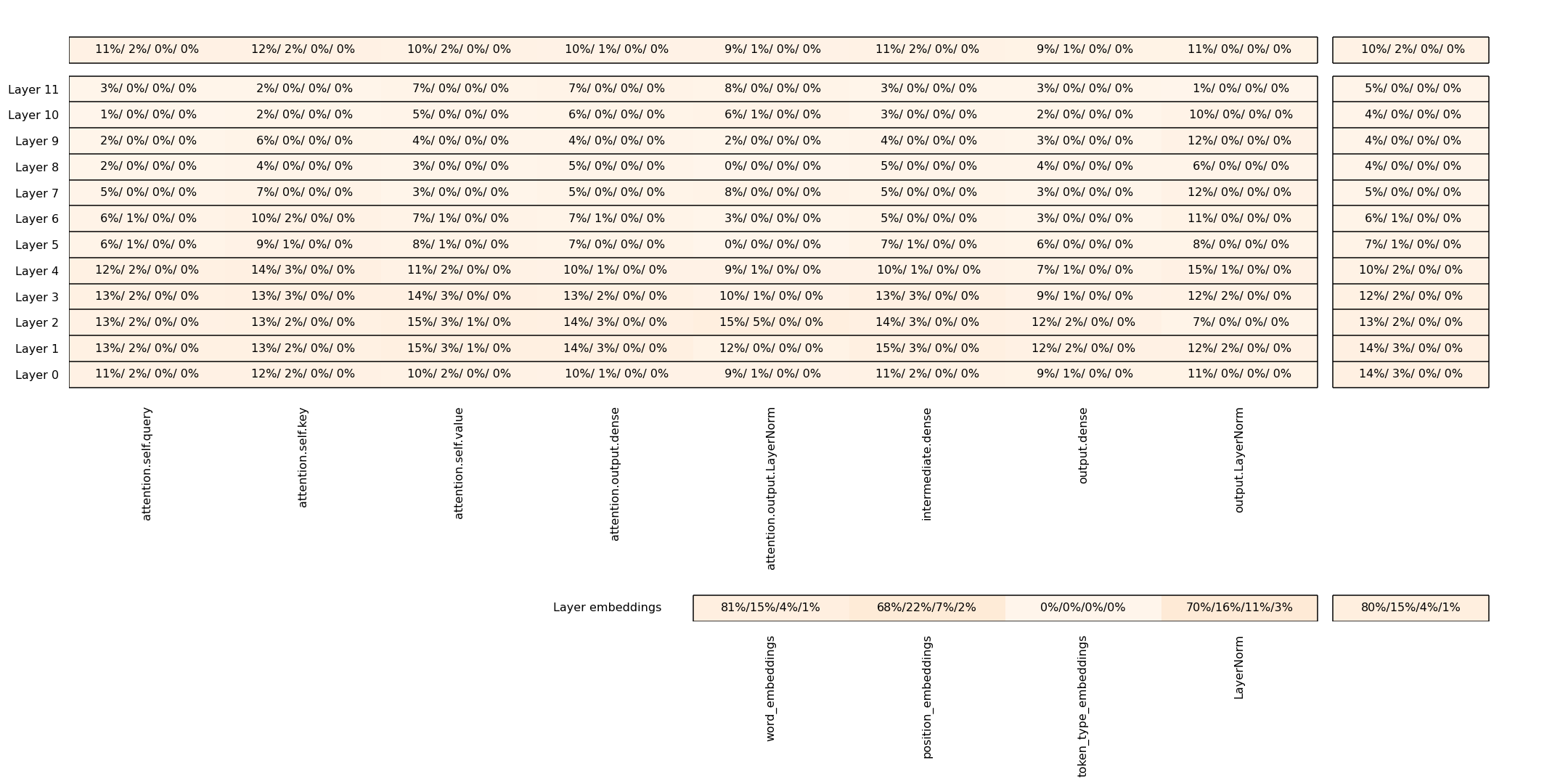}
  \caption{FCDL18 - Dialect}
  \end{subfigure}
  \caption{The percentage of common non-masked parameters for \modeldiffdeb across 5 runs with different initialization seeds. For each block, the numbers indicate the percentage of the number of common parameters across two, three, four, and five runs of a subnetwork, respectively.}
  \label{fig:overlap_diff}
\end{figure*}

Figures \ref{fig:overlap_mod}, \ref{fig:overlap_modseq} and \ref{fig:overlap_diff} show the percentage of common non-masked parameters in subnetworks on a particular weight matrix/layer across 5 runs for \modelmodularpar, \modelmodularseq, and \modeldiffdeb, respectively. We report the percentage of overlap between the subnetworks of two, three, four, and five runs, separated with the ``/''. The results show that the equivalent subnetworks across various runs (with different initialization seeds) seem to be largely separated. This results are consistent with the observations on the lottery ticket hypothesis on large neural networks~\cite{lottery_ticket_BERT_pruning_Chen2020}.

\end{document}